\documentclass[10pt]{article} 
\usepackage{hyperref}
\usepackage{url}
\usepackage{smile}
\usepackage{graphicx} 
\usepackage{algorithm}
\usepackage{algorithmic}
\usepackage{epstopdf}
\usepackage[margin=1.0in]{geometry}
\usepackage[export]{adjustbox}
\usepackage{mathtools, cuted}
\usepackage{bbm}
\usepackage{wrapfig}
\usepackage{subcaption}
\usepackage{caption}
\usepackage{enumitem}
\usepackage{tabularx}
\usepackage{smile}
\usepackage{tcolorbox}
\usepackage{enumitem}
\usepackage{mathtools, cuted}
\usepackage{caption} 
\usepackage{subcaption}

\numberwithin{equation}{section}

\renewcommand{\frac}{\tfrac}

\usepackage[nocompress]{cite}

\hypersetup{
    colorlinks=true,
    linkcolor=blue,
    anchorcolor=blue, 
    citecolor=blue,
}

\allowdisplaybreaks

\title{
 Policy Mirror Descent Inherently Explores Action Space
 \thanks{This research was partially supported by DMS-1953199 and AFOSR FA9550-22-1-0447.}
    }
\author{
    Yan Li   \thanks{H. Milton Stewart School of Industrial and Systems Engineering, Georgia Institute of Technology, Atlanta, GA, 30332. (E-mail: \url{yli939@gatech.edu}).}
 \and
    Guanghui Lan \thanks{H. Milton Stewart School of Industrial and Systems Engineering, Georgia Institute of Technology, Atlanta, GA, 30332. (E-mail: \url{george.lan@isye.gatech.edu}).}
}
\date{\vspace{-5ex}}

\begin{document}
{
\makeatletter
\addtocounter{footnote}{1} 
\renewcommand\thefootnote{\@fnsymbol\c@footnote}%
\makeatother
\maketitle
}

\maketitle

\begin{abstract}
Explicit exploration in the action space was assumed to be indispensable for online policy gradient methods to avoid a drastic degradation in sample complexity, for solving general reinforcement learning problems over finite state and action spaces. In this paper, we establish for the first time an $\tilde{\mathcal{O}}(1/\epsilon^2)$ sample complexity for online policy gradient methods without incorporating any exploration strategies. The essential development consists of two new on-policy evaluation operators and a novel analysis of the stochastic policy mirror descent method (SPMD) \cite{lan2022policy}. SPMD with the first evaluation operator, called value-based estimation,  tailors to the Kullback-Leibler divergence. Provided the Markov chains on the state space of generated policies are uniformly mixing with non-diminishing minimal visitation measure, an $\tilde{\mathcal{O}}(1/\epsilon^2)$ sample complexity is obtained with a linear dependence on the size of the action space. SPMD with the second evaluation operator, namely truncated on-policy Monte Carlo (TOMC), attains an $\tilde{\mathcal{O}}(\mathcal{H}_{\mathcal{D}}/\epsilon^2)$ sample complexity, where $\mathcal{H}_{\mathcal{D}}$ mildly depends on  the effective horizon and the size of the action space with properly chosen Bregman divergence (e.g., Tsallis divergence). SPMD with TOMC also exhibits stronger convergence properties in that it controls the optimality gap with high probability rather than in expectation. In contrast to explicit exploration, these new policy gradient methods can prevent repeatedly committing to potentially high-risk actions when searching for optimal policies. 
\end{abstract}


\section{Introduction}\label{sec_intro}

We consider a discrete time Markov decision process (MDP)  denoted by the quintuple $\cM = (\cS, \cA, \cP, c, \gamma)$, where $\cS$ denotes the finite state space, $\cA$ denotes the finite action space, $\cP: \cS \times \cS \times \cA \to [0, 1]$ specifies the transition kernel, $c: \cS \times \cA \to \RR$ denotes the cost function, and $\gamma \in (0,1)$ denotes the discount factor. 
We assume $0 \leq c(s,a) \leq 1$ for all $(s,a) \in \cS \times \cA$.

A randomized, stationary policy $\pi : \cS \to \Delta_{\cA}$ maps a given state $s\in \cS$ into $\pi(\cdot|s) \in \Delta_{\cA}$,
with $\Delta_{\cA}$ being the probability simplex over $\cA$.
The set of all such policies is denoted by $\Pi$.
A policy $\pi$ and a transition kernel $\cP$ jointly induce a stochastic process $\{(S_t, A_t)\}_{t \in \ZZ^*}$ on the state-action space $\cS \times \cA$, with $\ZZ^*$ denoting the set of nonnegative integers.
Specifically, at any timestep $t$,  the policy governs the action to be made given the current state $S_t$, by $A_t \sim \pi(\cdot| S_t)$. 
Then a cost $c(S_t, A_t)$ is incurred, followed by the state transition $S_{t+1} \sim \cP(\cdot|S_t, A_t)$. 
The decision process is repeated iteratively at future timesteps.

The performance of a policy $\pi $ is measured via its value function $V^{\pi}: \cS \to \RR$, defined as
\begin{align*}
V^{\pi} (s) \coloneqq \EE \sbr{\tsum_{t=0}^\infty \gamma^t c(S_t, A_t)  \big| S_0 = s, A_t \sim \pi(\cdot|S_t), S_{t+1} \sim \cP(\cdot|S_t,A_t)  }.
\end{align*}
The  planning of the MDP is to find an optimal policy $\pi^*$, with  
$V^{\pi^*}(s) \leq V^{\pi} (s)$  for all $ s \in \cS$ and $\pi \in \Pi$.
The optimal value function is defined as $V^* (s) \coloneqq \min_{\pi \in \Pi} V^{\pi}(s)$, for every $s \in \cS$.
Existence of an optimal $\pi^*$  is well-known in the literature of dynamic programming \cite{puterman2014markov}. 
With the existence of the $\pi^*$,  
one can instead solve the following single-objective optimization problem, 
\begin{align}
\textstyle
\label{eq:mdp_single_obj}
\min_{\pi} \cbr{f(\pi) \coloneqq \EE_{\vartheta} \sbr{V^{\pi}(s)}}, ~~ \mathrm{s.t.} ~~ \pi(\cdot|s) \in \Delta_{\cA}, \forall s \in \cS,
\end{align}
where $\vartheta \in \Delta_{\cS}$ is an arbitrarily chosen distribution on the state space\footnote{Indeed, the proposed method in this manuscript does not require information on $\vartheta$, and the results hold simultaneously for all $\vartheta \in \Delta_{\cS}$.}.
There has been a surge of interests in designing efficient first-order methods for directly searching the optimal policy \cite{agarwal2020optimality, cen2021fast, schulman2015trust, shani2020adaptive, liu2019neural}, despite the objective \eqref{eq:mdp_single_obj} being non-convex.
These methods utilize the first-order information of  \eqref{eq:mdp_single_obj} for policy improvement, and are hence termed policy gradient (PG) methods.
Notably, the key component for constructing the first-order information of a policy  is the so-called state-action value function, also known as the Q-function:
\begin{align}\label{def_q_function}
Q^{\pi} (s, a) \coloneqq \EE \sbr{\tsum_{t=0}^\infty \gamma^t  c(S_t, A_t) \big| S_0 = s, A_0 = a, A_t \sim \pi(\cdot|S_t), S_{t+1} \sim \cP(\cdot|S_t,A_t)  }.
\end{align}

When the model (i.e., $\cP$ and $c$) is known, $Q^{\pi}$ can be  computed exactly by solving a linear system, or approximately, with high accuracy,  after several  fixed-point iterations. 
In this case, both the projected policy gradient \cite{agarwal2020optimality} and the natural policy gradient  \cite{liu2019neural, agarwal2020optimality} converge at a sublinear rate when adopting constant stepsizes.
By carefully solving a sequence of entropy-regularized MDPs, with diminishing regularizations and increasing stepsizes, 
\cite{lan2022policy} proposes the first linearly-converging PG method for un-regularized MDPs. 
This is further simplified in \cite{xiao2022convergence, lan2022block, li2022first}, which drops the regularization while retaining the linear convergence.
Beyond the optimality gap, convergence of the policy has been studied in \cite{li2022homotopic}.
To cope with large state and action spaces, PG with function approximation has been discussed in \cite{liu2019neural, schulman2015trust, schulman2017proximal},
and more recently,~in \cite{lan2022policy_general}.

With an unknown model, most stochastic PG methods  can be categorized into the  actor-critic paradigm \cite{konda1999actor}. 
In a nutshell, these methods first perform policy evaluation to obtain a noisy estimator of the Q-function, 
using samples collected by interacting with the environment. 
Then stochastic first-order information can be constructed for updating the policy in the policy improvement step.
The sample complexity of stochastic PG methods has been the primary concern, as it takes up the majority of computing budget. 

Based on the capability for trajectory generation, we categorize policy evaluation operators into online and generator-based variants. 
For generator-based policy evaluation operators, one can arbitrarily choose a starting state-action pair when generating a trajectory, and is capable of restarting the trajectory at any time. 
Applications of these operators are typically limited to training in a simulated environment.
Online policy evaluation operators are considered to be more realistic, 
where one simply follows the current policy without assuming the power of choosing the starting state-action pair, or any form of restarting capability. We term stochastic PG methods with generator-based policy evaluation operators as generator-based PG methods,
and similarly define online PG methods.

Sample complexities of generator-based PG methods are relatively well understood. 
In particular, 
$\tilde{\cO}(1/\epsilon^3)$ sample complexity has been established \cite{shani2020adaptive} for a stochastic variant of natural policy gradient. 
This has~been recently improved to $\tilde{\cO}(1/\epsilon^2)$ in \cite{lan2022policy},
attaining  the optimal dependence on the  target accuracy.
 The proposed method therein is further extended in \cite{li2022homotopic, xiao2022convergence}, with a simplified analysis. 
%


The situation becomes considerably more involved for online PG methods.
In particular, as the policy progresses towards the optimal policy, non-optimal actions get rarely explored within policy evaluation given their diminishing policy values (i.e., probabilities assigned by the policy).
Consequently, first-order information for these actions becomes increasingly difficult to obtain.
Below, we review a few prior approaches that aim to address this lack of exploration.

The first approach, and widely adopted in literature, assumes the policy will take each action with a probability bounded away from zero \cite{lan2022policy, alacaoglu2022natural, abbasi2019politex}.
With this assumption, numerous online policy evaluation methods, including the celebrated TD learning \cite{sutton1988learning, tsitsiklis1996analysis, kotsalis2022simple}, can be applied, and a sample complexity of $\tilde{\cO}(1/\epsilon^2)$ is attained for finding an $\epsilon$-optimal policy \cite{lan2022policy, alacaoglu2022natural}.
A key limitation of this assumption is its contradiction with 
the goal of planning, which seeks to identify which actions to  avoid (assigning zero probability).
Consequently, this assumption does not hold whenever the policy is close to the set of optimal policies. 

The second approach considers forcing explicit exploration over the action space to ensure every action is sampled with non-zero probability.
Above all,  a commonly adopted technique is the so-called $\epsilon$-exploration, where the policy is mixed with the uniform distribution over the action space,  within policy evaluation \cite{alacaoglu2022natural} (resp. policy improvement, \cite{khodadadian2022finite}), for which a sample complexity of $\tilde{\cO}(1/\epsilon^4)$ (resp. $\tilde{\cO}(1/\epsilon^6)$) is obtained. 
A more delicate method, based on policy perturbation, that periodically switches the policy to an exploring one within policy evaluation \cite{li2022stochastic},  attains an $\tilde{\cO}(1/\epsilon^2)$ sample complexity. 
We refer to this type of explicit exploration  as myopic exploration, since it ignores the cause for a particular action to have small probability. 
In particular, sub-optimal, and potentially high-risk actions that have been identified by policy optimization will be repeatedly taken within policy evaluation. 

The third approach,  taken in \cite{hu2022actorcritic, liu2019neural},  avoids myopic exploration. Rather, it adopts the simplest form of actor-critic methods without any explicit exploration or intervention (e.g., restart).
The global convergence exploits the weighted convergence of TD method for estimating the Q-function,
where the weights of state-action pairs are determined by the visitation measure of the current policy \cite{bhandari2018finite, telgarsky2022stochastic}.
The simplicity of this approach, on the other hand, seems to come with a steep price. 
In particular, an $\tilde{\cO}(1/\epsilon^{16})$ (resp. $\tilde{\cO}(1/\epsilon^8)$) sample complexity is established in \cite{hu2022actorcritic}\footnote{It seems the sample complexity of \cite{hu2022actorcritic} can be potentially strengthened to $\tilde{\cO}(1/\epsilon^6)$ by utilizing results in \cite{telgarsky2022stochastic}.
} (resp. \cite{liu2019neural}).

\vspace{0.05in}
{\bf Contributions.} This manuscript proposes an online policy gradient method with no explicit exploration, 
which  consequently enjoys implementation simplicity and avoids the pitfall of myopic exploration.
More importantly, it does so while attaining an optimal sample complexity $\tilde{\cO}(1/\epsilon^2)$ in terms of its dependence on the target accuracy.
In particular,  we summarize the main contributions as follows. 

First, we present two  online evaluation  operators that build upon the simple on-policy Monte Carlo (OMC) method.
The first evaluation operator, named value-based estimation (VBE), adopts a conceptually two-step process and has two variations.
VBE-I maintains two independent trajectories.  An estimator of the value function is constructed by using one trajectory, which is then used in conjunction with the other trajectory to construct an estimator of the Q-function.
 VBE-II, on the other hand,   only requires a single trajectory.
Notably, the bias for each state-action pair in the VBE estimator diminishes
with a rate that is linear in the trajectory length.
The second operator, named truncated on-policy Monte Carlo (TOMC), maintains a single trajectory, and 
deviates from its OMC counterpart by a simple truncation step that returns a trivial upper bound on the true value of Q-function, whenever the action has  a low policy value. 
Notably, as an evaluation operator, TOMC does not forcefully seek uniform control of the bias for every~action.

Second, we revisit a first-order policy optimization method, named stochastic policy mirror descent (SPMD, \cite{lan2022policy}), 
which performs mirror descent type policy update with stochastic first-order information.
The only assumption made herein is the state chains of policies generated by SPMD being uniformly mixing and exploring.
No exploration assumption on the action space, of any form, is made, as opposed to a~large body of literature for actor-critic methods, including the original development of SPMD. 
By adopting VBE operator for policy evaluation, 
 we establish the global convergence of SPMD instantiated with the Kullback-Leibler (KL) divergence.
In particular, we establish   
an $\tilde{\cO}(1/\epsilon^2)$ sample complexity,  for which the expected optimality gap of the best-iterate policy falls below $\epsilon$.

Third, we show that 
under the same assumption of the state chains, 
using a single trajectory of sufficient length,
 SPMD with the TOMC operator exhibits inherent exploration over the action space, in the sense that policy value is lower bounded for every optimal action with high probability.
We then establish a novel bound on the accumulated noise of SPMD, with a probabilistic argument of potentially independent interest, from which an $\tilde{\cO}(\cH_{\cD}/\epsilon^2)$ sample complexity is obtained. 
Here $\cH_{\cD}$ is a divergence-dependent function of the effective horizon and the size of the action space. 
We show different Bregman divergences lead to drastically different $\cH_{\cD}$.
In particular,  KL divergence yields an exponential dependence on the effective horizon,
while divergence induced by the negative Tsallis entropy yields a polynomial dependence in both the effective horizon and the action space.
For the latter divergence, we provide a simple bisection-based subroutine for solving the proximal policy update subproblem with linear convergence.
Notably, SPMD with TOMC operator directly controls the optimality gap of the best-iterate policy in high probability, which is stronger than the expectation bound associated with the VBE operator.

\subsection{Notation and Terminology}
For any policy $\pi$, we define the discounted state visitation measure 
$d_{s}^\pi(s') \coloneqq (1-\gamma) \sum_{t=0}^\infty \gamma^t \PP^{\pi}(S_t = s' | S_0 = s)$, where $\PP^{\pi}(S_t = s'|S_0 = s)$ denotes the probability of reaching state $s'$ at time $t$ by following policy $\pi$,  when starting at state $s$ at time $0$.
Accordingly, we define $d_{\vartheta}^{\pi} (\cdot) \coloneqq \EE_{s \sim \vartheta} d_{s}^{\pi}(\cdot)$ for any $\vartheta \in \Delta_{\cS}$.
We denote the stationary state distribution of policy $\pi$ by $\nu^{\pi}$, which satisfies $\nu^\pi(s) = \sum_{s' \in \cS} \nu^{\pi}(s') \PP^{\pi}(s|s')$ where $\PP^{\pi}(s|s') \coloneqq  \sum_{a \in \cA} \pi(a|s') \cP(s|s', a)$.
  We use  $\Pi^*$ to denote  the set of optimal stationary policies, and identify $\cZ \equiv \cS \times \cA$ as the state-action space.
For any set $\cX$, we denote $\mathrm{ReInt}(\cX)$ as the relative interior of a set $\cX$, and denote $\mathrm{ReBd}(\cX)$ as the relative boundary.
Accordingly, $\mathrm{ReInt}(\Pi)$ denotes the set of policies that assign a positive probability to every action at every state.

For a strictly convex function  $w$ with domain containing $\Delta_{\cA}$, we define 
\begin{align}
\label{def:kl_bregman}
D^{\pi}_{\pi'} (s) \coloneqq w(\pi(\cdot|s)) - w(\pi'(\cdot|s)) - \inner{\nabla w(\pi'(\cdot|s))}{\pi(\cdot|s) - \pi'(\cdot|s) },
\end{align}
where $\nabla w(\pi'(\cdot|s))$ denotes the subgradient of $w(\pi'(\cdot|s))$ w.r.t. $\pi'(\cdot|s)$. 
This corresponds to the Bregman divergence applied to the policy at state $s \in \cS$, with the distance-generating function being $w$.

\section{Stochastic Policy Mirror Descent}\label{spmd_technical_background}

We start this section by briefly reviewing the stochastic policy mirror descent method (SPMD),  proposed in \cite{lan2022policy}.
Then the tension between policy optimization and evaluation is introduced. We discuss related works together with their limitations, 
which serves as the motivation of our technical developments in the ensuing sections.

Each iteration of SPMD consists of the following update of the policy:
\begin{align}
\label{spmd_update}
\textstyle
\pi_{k+1}(\cdot | s) = \argmin_{p(\cdot |s) \in \Delta_{\cA}} \eta_k \inner{Q^{\pi_k, \xi_k}(s, \cdot)}{p(\cdot|s)}  + D^{p}_{\pi_k}(s), ~~ \forall s \in \cS,
\end{align}
where $Q^{\pi_k, \xi_k}$ denotes the stochastic estimator of $Q^{\pi_k}$, and  $\xi_k$ denotes the random variables used for the construction of the estimator.
Throughout the rest of our discussions, we define 
\begin{align}
\label{def_delta_noise}
\delta_t \coloneqq Q^{\pi_t} -  Q^{\pi_t, \xi_t}
\end{align}
as the noise in the stochastic estimator. It is worth noting that the above definition is  asymmetric with respect to the position of $Q^{\pi_t}$ and $Q^{\pi_t, \xi_t}$.
This would be particularly helpful for our discussions in Section \ref{sec_spmd_omc}.



\vspace{0.05in}
{\bf Interplay between policy optimization and evaluation.}
A large body of literature assumes uniform control on the noise in $Q^{\pi_k, \xi_k}$. Namely, 
\begin{align}\label{eq_unif_nose_condition}
\EE_{\xi_k} \norm{Q^{\pi_k, \xi_k} - Q^{\pi_k}}_\infty^2 \leq \sigma_k^2, ~\text{or}~
\norm{\EE_{\xi_k} Q^{\pi_k, \xi_k} - Q^{\pi}}_\infty \leq \varepsilon_k, 
\end{align}
for pre-specified target noise level $\sigma_k$ (or $\varepsilon_k$) that diminishes to zero  \cite{lan2022policy, xiao2022convergence}. 
This can be readily satisfied if a generator/simulator can be accessed, an assumption that clearly limits the applicability of stochastic PG methods.
A more appealing alternative is to apply online policy evaluation operators that simply follows the current policy.
TD type evaluation operators
  are arguably the most studied option in this scenario. 
However, to output an estimate satisfying \eqref{eq_unif_nose_condition}, it requires,
at minimal, that $\min_{a \in \cA} \pi_k(a|s) $  being bounded away from zero \cite{tsitsiklis1996analysis, kotsalis2022simple} for every state $s\in \cS$. 
This condition breaks eventually, as the policy progresses to $\Pi^*$. 
To illustrate, we recall the well-known characterization of optimal stationary policies. 

\begin{lemma}[Lemma 4, \cite{li2022homotopic}]\label{lemma:optimal_policy_set}
 The set of optimal stationary policies $\Pi^*$ is given by
\begin{align*}
\textstyle
\Pi^* = \cbr{
\pi \in \Pi: \mathrm{supp}(\pi(\cdot|s)) \subseteq \cA^*_s \coloneqq \Argmin_{a \in \cA} Q^*(s, a), ~ \forall s \in \cS
},
\end{align*}
where $Q^*(s,a) \coloneqq \min_{\pi \in \Pi} Q^{\pi}(s,a)$ denotes the optimal Q-function, and $\mathrm{supp}(p)$ denotes the support of distribution $p \in \Delta_{\cA}$.
\end{lemma}

In view of Lemma \ref{lemma:optimal_policy_set}, if $\cA^*_s = \cA$ for every $s \in \cS$, then every policy is optimal and consequently the MDP is trivial.
For any nontrivial MDP instance, 
one can make the following observation for online policy evaluation operators: 
\begin{align}\label{tension_observation}
\text{If} ~ \pi_k \to \Pi^* ~ \Rightarrow
~
\pi_k(a|s) \to 0, ~ \text{for every} ~  a \notin \cA^*_s 
~ 
\Rightarrow ~
\eqref{eq_unif_nose_condition}~  \text{fails for $k$ large enough}.
\end{align}

As we have discussed in Section \ref{sec_intro},
there have been two main approaches in the literature of online PG methods for addressing \eqref{tension_observation}.
One approach forces \eqref{eq_unif_nose_condition} with myopic exploration, with either $\epsilon$-exploration \cite{alacaoglu2022natural, khodadadian2022finite}, or policy perturbation \cite{li2022stochastic}.
A drawback of myopic exploration, aside from another layer of complexity in implementation,  is its repeating sample of non-optimal or high-risk actions, even if the policy has identified these actions. 
Another approach makes no changes to the stochastic PG methods, as it does not require uniform control over the noise \eqref{eq_unif_nose_condition}. Instead, weighted convergence of TD are utilized in the analysis, resulting in a worse sample complexity \cite{hu2022actorcritic, liu2019neural} compared to the myopic (explicit) exploration. 

\vspace{0.05in}
{\bf Exploration in bandit/MDP.}
It would be remiss if this manuscript focuses on exploration without discussing decades of development in the bandit literature.
This type of problems can be viewed as solving a single-state MDP with unknown cost function.
Each round of interaction with the environment returns the cost information for the action taken in that round, and consequently solving this problem requires a proper tradeoff between exploitation and exploration. 
A general rule of algorithm design in this domain is optimism in the face of uncertainty,
which, above all, motivates the celebrated UCB method \cite{lai1985asymptotically}.
Another related method that balances exploration and exploitation is EXP3 \cite{auer2002nonstochastic}, which can be viewed as an instance of stochastic mirror descent \cite{nemirovskij1983problem, nemirovski2009robust} instantiated with the Kullback-Leibler divergence.

Exploration techniques based on the principle of optimism have also been extended to MDPs,
which can efficiently promote active exploration in both state and action spaces.
Many of these methods are inherently model-based \cite{auer2006logarithmic, azar2017minimax, auer2008near}, as they either construct a confidence set of models and perform optimistic planning therein, 
or require estimating quantities of the same dimension as the model \cite{cai2020provably}.
It seems that if one seeks to avoid estimating the model or quantities of similar dimension, 
 much attention has been devoted to value-based methods (e.g., bonus-based Q-learning \cite{strehl2006pac, jin2018q}).
Some prior development have also proposed model-free actor-critic methods aiming for efficient exploration \cite{agarwal2020pc,feng2021provably}.
  These methods in turn require storing historical policies, leading to a memory footprint that exceeds model-based methods. 
  The obtained sample complexities therein also have non-optimal dependence on the target accuracy.

\section{On-policy Monte-Carlo}\label{sec_on_policy_eval}



This section describes a simple online policy evaluation method, named on-policy Monte-Carlo (OMC), that serves as the base method for constructing the policy evaluation operators to be studied in the following sections. 
OMC can be applied to estimate either the Q-function or the value function of a fixed policy, which collects samples by following the policy without any intervention.

 For any to-be-evaluated policy $\pi$, 
 let $\cbr{S_t^{\pi}}_{t \in \ZZ^*} \subseteq \cS $ be the Markov chain generated by policy $\pi$.
  We often omit the index set $\ZZ^*$ when the context is clear, for notational simplicity.
 Suppose  $\cbr{S_t^{\pi}}_{t \in \ZZ^*}  $ is irreducible and aperiodic,
then it is clear that $\cbr{S_t} \coloneqq \cbr{S_t^\pi}$ is geometrically mixing \cite{levin2017markov}, with
\begin{align}\label{eq_fast_mixing_state}
\tsum_{s\in \cS} \abs{\PP(S_t = s| S_0 = s_0)  - \nu^\pi(s)  } \leq C \rho^{t+1}, ~ C >0, ~\rho \in (0,1),
\end{align}
for any $s_0 \in \cS$.
Let $\nu^\pi$ denotes its stationary distribution.
 Consequently, it is straightforward to verify that 
Markov chain $\cbr{(S_t, A_t)} \subseteq \cZ$ induced by the policy over the state-action space, where $A_t \sim \pi(\cdot| S_t)$,  is also geometrically mixing.
That is, for any $(S_0 ,A_0) \in \cZ$, 
\begin{align}\label{eq_fast_mixing}
\tsum_{s\in \cS, a \in \cA} \abs{\PP(S_t = s, A_t = a | S_0, A_0)  - \nu^\pi(s) \pi(a|s) } \leq C \rho^t, ~ C>0, ~\rho \in (0,1),
\end{align}
Let $\cbr{Z_t}  \coloneqq \cbr{(S_t, A_t)}$, 
the stationary distribution of $\cbr{Z_t}$ is then given by 
\begin{align}\label{def_stationary_state_action_chain}
\sigma^\pi(s,a)  = \nu^\pi (s)  \pi(a|s).
\end{align} 

Throughout the rest of our discussion, we make only the assumption that the chain $\cbr{S^{\pi_k}_t}$ is uniformly mixing and exploring over the state space, for each of the policies $\cbr{\pi_k}$ generated by SPMD.

\begin{assumption}\label{assump_unif_mixing}
Let $\cbr{\pi_i}_{i \leq k}$ be policies generated by running SPMD for $k$ iterations. There exists a common $(C, \rho)$, such that \eqref{eq_fast_mixing_state} (and consequently \eqref{eq_fast_mixing}) holds for all $\pi_i$ with $i \leq k$,
In addition, there exists an $\underline{\nu} > 0$, such that $\min_{i \leq k, s\in \cS} \nu^{\pi_i}(s) \geq \underline{\nu}$.
\end{assumption}


Assumption \ref{assump_unif_mixing} can be satisfied, for example, when the underlying MDP is ergodic.
It can be further relaxed for the purpose of our ensuing discussion.
Indeed, a major part of our technical development, in particular, Section \ref{sec_spmd_omc}, only requires \eqref{eq_fast_mixing_state} and $\nu^{\pi^*} \succeq \mathbf{1} \underline{\nu} \succ 0$ to hold for some  deterministic optimal policy $\pi^*$ (see Remark \ref{remark_relax_assump}).
Nevertheless, it should be noted that Assumption \ref{assump_unif_mixing} is, by no means, a weak assumption,
as it implicitly assumes exploration over the state space. 
As we have mentioned in Section \ref{spmd_technical_background}, 
exploration over the state space of  MDPs has been an active research area, especially in regret analysis.
However, theses methods seem to be either model- or value-based, or with a sample complexity that has non-optimal dependence on the target accuracy.



\subsection{On-policy Monte Carlo for Q-function}\label{sec_omc_eval}

We now proceed to describe the on-policy Monte Carlo (OMC) method  for estimating the state-action value function $Q^{\pi}$ defined in \eqref{def_q_function}.

For any $z \equiv (s,a) \in \cZ$ and any $n \geq 1$, we define $\cT_n(z) \coloneqq  \cbr{t \leq n-1: Z_t = z} $, and 
\begin{align*}
\tau(z) \coloneqq \begin{cases}
 \min \cT_n(z), & ~ \cT_n(z) \neq \emptyset, \\
 n, & ~ \cT_n(z)  = \emptyset.
 \end{cases}
\end{align*}
By definition, if the chain of state-action pair $\cbr{Z_t}$ generated by the policy  reaches the target state-action pair $z \in \cZ$ before $(n-1)$-th timestep, then $\tau(z)$ is the first timestep that $\cbr{Z_t}$ reaches $z$, otherwise, $\tau(z)$ takes the value of $n$.
Clearly, $\tau(z)$ is a stopping time of $\cbr{Z_t}$. 

Now consider the following random variable: 
\begin{align}\label{omc_estimate}
\hat{Q}^{\pi}(s,a) \coloneqq \begin{cases}  \tsum_{t = \tau(z)}^{n-1} \gamma^{t - \tau(z)} c(S_t, A_t),  & \tau(z) \leq n-1, \\
0, & \tau(z) = n.
\end{cases}
\end{align}
The next lemma characterizes the bias of  $\hat{Q}^{\pi}(s,a)$. 
\begin{lemma}\label{lemma_bias_omc}
Let $(C, \rho)$ be defined as in Assumption \ref{assump_unif_mixing}, and $\sigma^\pi$ be defined as in \eqref{def_stationary_state_action_chain}.
For any $z \equiv (s,a) \in \cS \times \cA$ with $\sigma^{\pi}(z) > 0$, let $t_{\mathrm{mix}} (z) =  \ceil{\log_\rho ( \tfrac{\sigma^\pi(z)}{2 C})}$.
Then we have 
\begin{align*}
\abs{ \EE \sbr{\hat{Q}^{\pi}(s,a) - Q^{\pi}(s,a)} } 
\leq 
\tfrac{2(n+1)}{1-\gamma} \sbr{ \gamma^{n-1} + (1- \tfrac{\sigma^\pi(z)}{2 t_{\mathrm{mix}(z)}})^{n-1}}.
\end{align*}
\end{lemma}

\begin{proof}
Since $\tau(z)$ is a stopping time of $\cbr{Z_t}$, given the strong Markov property,  
\begin{align*}
\abs{ \EE \sbr{\hat{Q}^{\pi}(s,a) - Q^{\pi}(s,a)} } 
& \leq \abs{Q^{\pi}(s,a)} \PP(\tau(z) = n) 
+ \abs{\tsum_{i=0}^{n-1} \EE \sbr{\hat{Q}^{\pi}(s,a) - Q^{\pi}(s,a) | \tau(z) = i } \cdot \PP(\tau(z) = i) }  \\
& \leq 
\tfrac{1}{1-\gamma} \PP(\tau(z) = n)
+ \tsum_{i=0}^{n-1} \abs{\EE \sbr{\hat{Q}^{\pi}(s,a) - Q^{\pi}(s,a) | \tau(z) = i } } \cdot \PP(\tau(z) = i) \\
& \leq 
\tfrac{1}{1-\gamma} \PP(\tau(z) = n)
+ \tfrac{1}{1-\gamma} \tsum_{i=0}^{n-1} \gamma^{n - i} \PP(\tau(z) = i)  \\
& = \tfrac{1}{1-\gamma} 
\EE \sbr{ \gamma^{n - \tau(z)}}.
\end{align*}


For any $z \in \cZ$ with $\sigma^{\pi}(z) > 0$. 
If $Z_0 = z$, then $\tau(z) = 0$ and hence $\EE \sbr{ \gamma^{n - \tau(z)}} = \gamma^n$.
Let us consider the scenario where $Z_0 \neq z$. 
Define $t_{\mathrm{mix}} (z) = \ceil{\log_\rho ( \tfrac{\sigma^\pi(z)}{2 C})}$, which is abbreviated as $t_{\mathrm{mix}}$ when the context is clear. 
For any $1 \leq i \leq n-1$, 
define random variables $Y_k \coloneqq Z_{k \cdot t_{\mathrm{mix}}}$ for $0 \leq k \leq  \overline{k} \equiv \ceil{i / t_{\mathrm{mix}}} -1$. 
Then 
\begin{align}
\PP(\tau(z) = i | Z_0 ) 
& = \PP(Z_t \neq z, \forall t < i, Z_i = z | Z_0) \nonumber \\
& \leq 
\PP(Y_k \neq z, \forall k \leq \overline{k}, Z_i = z | Z_0) \nonumber\\
& = 
\tsum_{y_k \neq z,  0 \leq k \leq \overline{k} } \PP(Z_i = z, Y_k = y_k , \forall k \leq \overline{k} | Z_0)\nonumber \\ 
&\overset{(a)}{=} 
\tsum_{y_k \neq z,  0 \leq k \leq \overline{k} }
\PP(Z_i = z | Y_{\overline{k}} = y_{\overline{k}})  
\prod_{k=1}^{\overline{k}} \PP(Y_k = y_k | Y_{k-1} = y_{k-1} ) \nonumber \\
& \leq \tsum_{y_k \neq z,  0 \leq k \leq \overline{k} } 
\prod_{k=1}^{\overline{k}} \PP(Y_k = y_k | Y_{k-1} = y_{k-1} ) \nonumber \\
& \overset{(b)}{\leq} 
( 1- \tfrac{\sigma^\pi(z)}{2})^{ \ceil{i / t_{\mathrm{mix}}} -1} ,
\label{prob_for_hitting_time_z}
\end{align}
where $(a)$ follows from the Markov property, and $(b)$ follows from that for any $y_{k-1} \in \cZ$ and any $1 \leq k \leq \overline{k}$, 
\begin{align*}
\tsum_{y_k \neq z} \PP(Y_k = y_k | Y_{k-1} = y_{k-1}) & = 1 -  \PP(Y_k = z | Y_{k-1} = y_{k-1})  \\
 & = 1 -  \PP(Z_{k \cdot t_{\mathrm{mix}}} = z |Z_{(k-1) \cdot t_{\mathrm{mix}}} = y_{k-1}) \\
& \leq 1 -  \tfrac{\sigma^\pi(z)}{2},
\end{align*}
given the definition of $\cbr{Y_k}$,  inequality \eqref{eq_fast_mixing},  together with the definition of $t_{\mathrm{mix}}$.
With similar arguments, we can also show 
\begin{align*}
\PP(\tau(z) = n | Z_0 ) \leq
( 1- \tfrac{\sigma^\pi(z)}{2})^{ \ceil{n / t_{\mathrm{mix}}}  - 1}.
\end{align*}
Thus for $Z_0 \neq z$, we have 
\begin{align*}
\EE \sbr{ \gamma^{n - \tau(z)} }
&\leq \sbr{ \tsum_{i=1}^{n-1} ( 1- \tfrac{\sigma^\pi(z)}{2})^{ \ceil{i / t_{\mathrm{mix}}} -1} \gamma^{-i} 
+ ( 1- \tfrac{\sigma^\pi(z)}{2})^{ \ceil{n / t_{\mathrm{mix}}} -1} \gamma^{-n} } \gamma^n \\
& \leq 
2 \sbr{ \gamma^n \tsum_{i=0}^{n-1} ( 1- \tfrac{\sigma^\pi(z)}{2})^{ \tfrac{i}{ t_{\mathrm{mix}}} } \gamma^{-i} 
 + ( 1- \tfrac{\sigma^\pi(z)}{2})^{ \tfrac{n}{ t_{\mathrm{mix}}} } }\\
 & \overset{(a)}{\leq} 
2 \gamma^n  \tsum_{i=0}^{n-1} ( 1- \tfrac{\sigma^\pi(z)}{2 t_{\mathrm{mix}}})^i \gamma^{-i}
  + 2 ( 1- \tfrac{\sigma^\pi(z)}{2 t_{\mathrm{mix}}})^n,
\end{align*}
where 
$(a)$ follows from $1 - {\sigma^{\pi}(z)}/{2} \geq {1}/{2}$, and
$(b)$ follows from Bernoulli's inequality.
Now if $\gamma =  1- \tfrac{\sigma^\pi(z)}{2 t_{\mathrm{mix}}}$, then
$
\EE \sbr{ \gamma^{n - \tau(z)} } \leq (n+1) \gamma^n.
$
If $\gamma \neq 1- \tfrac{\sigma^\pi(z)}{2 t_{\mathrm{mix}}}$, then letting $ p_z =  \tfrac{\sigma^\pi(z)}{2 t_{\mathrm{mix}}}$, we obtain 
\begin{align*}
\EE \sbr{ \gamma^{n - \tau(z)} }
& \leq \tfrac{\gamma^n - (1-p_z)^n}{\gamma - (1-p_z)} + (1 - p_z)^n \\
& \overset{(a')}{\leq} 
n \max \cbr{
\gamma^{n-1}, (1- p_z)^{n-1}
}
+ (1 - p_z)^n \\
& \leq 
n \gamma^{n-1} 
+ (n+1) (1-p_z)^{n-1},
\end{align*}
where $(a')$ follows from the mean value theorem.

In conclusion, we obtain that for any $Z_0 \in \cZ$, and any $z \in \cZ$,
\begin{align*}
\EE \sbr{ \gamma^{n - \tau(z)} } 
\leq 2 (n + 1) \sbr{ \gamma^{n-1} + (1- \tfrac{\sigma^\pi(z)}{2 t_{\mathrm{mix}}(z)})^{n-1}},
\end{align*}
from which the desired claim follows immediately.
\end{proof}


Unless stated otherwise, in this manuscript, a trajectory $\xi$ of length $n$ takes the form of $ \cbr{Z_0, \ldots, Z_{n-1}}$.
The $\mathrm{OMC}( n; m)$ procedure is described as follows. 
For each $1 \leq l \leq m$, we generate a trajectory of length $n$ starting from arbitrary $Z_0^{(l)} \in \cZ$, which is denoted by $\xi^{(l)} $.
Then we construct each $\hat{Q}^\pi_{(l)}(s,a)$ for any $(s,a) \in \cS \times \cA$,  defined as in \eqref{omc_estimate}.
Let $\xi = \cbr{\xi^{(l)}}_{1 \leq l \leq m}$ denote the collection of $m$ trajectories, the final estimator is given by 
\begin{align}\label{omc_estimate_batch}
Q^{\pi, \xi, (n, m)}_{\mathrm{OMC}}(s,a) \coloneqq \tfrac{1}{m} \tsum_{l=1}^m \hat{Q}^\pi_{(l)}(s,a), ~ \forall (s,a) \in \cS \times \cA.
\end{align}

It  should be noted that in the OMC estimator defined above, 
even with multiple trajectories ($m> 1$), 
the set of initial state-action pairs can be arbitrary.
This should be contrasted with policy  evaluation with a generative model, where one enumerates over all $(s,a) \in \cZ$, and generate a trajectory for each  pair.

\subsection{On-policy Monte Carlo for Value Function}

The OMC method discussed in Section \ref{sec_omc_eval}  can also be applied to estimate the value function $V^{\pi}$ for any policy $\pi$. 
Specifically, with a slight overloading of notations, for any $s \in \cS $ and any $n \geq 1$, we define $\cT_n(s) \coloneqq  \cbr{t \leq n-1: S_t = s} $, and 
\begin{align*}
\tau (s) \coloneqq \begin{cases}
 \min \cT_n(s), & ~ \cT_n(s) \neq \emptyset, \\
 n, & ~ \cT_n(s)  = \emptyset.
 \end{cases}
\end{align*}
Accordingly, let us define 
\begin{align}\label{omc_value_estimator_single}
\hat{V}^{\pi}(s) \coloneqq \begin{cases}  \tsum_{t = \tau(s)}^{n-1} \gamma^{t - \tau(s)} c(S_t, A_t),  & \tau(s) \leq n-1, \\
0, & \tau (s) = n.
\end{cases}
\end{align}
With the same argument as in Lemma \ref{lemma_bias_omc}, we can show fast bias reduction of the estimator above.
\begin{lemma}\label{lemma_omc_value_est_bias}
For any $s \in \cS$ with $\nu^{\pi}(s) > 0$, let $t_{\mathrm{mix}} (s) = \ceil{\log_\rho ( \tfrac{\nu^\pi(s)}{2 C})}$.
Then we have 
\begin{align*}
\abs{ \EE \sbr{\hat{V}^{\pi}(s) - V^{\pi}(s)} } 
\leq 
\tfrac{2(n+1)}{1-\gamma} \sbr{ \gamma^{n-1} + (1- \tfrac{\nu^\pi(s)}{2 t_{\mathrm{mix}(s)}})^{n-1}}.
\end{align*}
\end{lemma}  

Let $\xi = \cbr{\xi^{(l)}}_{1 \leq l \leq m}$ denote the collection of trajectories, 
each of length $n$, 
then the $\mathrm{OMC}(n;m)$ estimator  for $V^{\pi}$ is defined as 
\begin{align}\label{OMC_for_value}
V^{\pi, \xi, (n, m)}_{\mathrm{OMC}}(s) \coloneqq \tfrac{1}{m} \tsum_{l=1}^m \hat{V}^\pi_{(l)}(s), ~ \forall s \in \cS,
\end{align}
with each $\hat{V}^\pi_{(l)}$ constructed as in \eqref{omc_value_estimator_single} using trajectory $\xi^{(l)}$.


Before we conclude this section, we remark that on-policy Monte Carlo, as a simple policy evaluation method, has received attention in the literature \cite{singh1996reinforcement}.
On the other hand,  most prior discussions consider applying this estimation method within approximate dynamic programming,
 which performs greedy policy improvement step based on the estimator.
It seems that to attain finite-time convergence, this requires using a large $m$ to control the noise of the estimator, otherwise only asymptotic convergence has been established \cite{tsitsiklis2002convergence}.
More importantly, such an approach also requires explicit exploration over the action space. 

\begin{remark}\label{remark_vanilla_omc}
In view of Lemma \ref{lemma_bias_omc}, the OMC estimator \eqref{omc_estimate_batch}  for Q-function  still requires the to-be-evaluated policy to have a non-diminishing policy value for every action, in order to uniformly control the bias of the estimate.
This presumption breaks eventually given observation \eqref{tension_observation}.
On the other hand,  estimator \eqref{OMC_for_value} only estimates the value function, and it is not immediate how one could in turn obtain a quality estimator of the Q-function.
Consequently,  OMC estimators constructed in this section need further modifications before being incorporated into SPMD. 
\end{remark}

In the following sections, we introduce such modifications, by  constructing two  online policy evaluation operators that properly adapt and incorporate the OMC estimators \eqref{omc_estimate_batch} (or \eqref{OMC_for_value}).
More importantly, we establish that SPMD with these evaluation operators 
can perform efficient policy optimization with a single trajectory, while not employing any explicit exploration over the action space.



\section{SPMD with Value-based Estimation}\label{sec_spmd_vbe}

We begin this section by introducing the first policy evaluation operator, namely value-based estimation (VBE), for constructing the stochastic estimators $\{Q^{\pi_t, \xi_t}\}$ used in the policy update \eqref{spmd_update}.
We then proceed to establish the global convergence of SPMD with the VBE operator, and determine its sample complexity. 

\subsection{Value-based Estimation}
The construction of VBE
avoids direct estimation of the Q-function. 
Conceptually, it
performs a two-step process.
The first step estimates the value function of the policy,  
 which is utilized to construct the estimated Q-function in the second step.
Following this general recipe, we first describe  VBE-I, which maintains two independent trajectories, with one used in estimating the value function,
and the other used in the construction of the Q-function.
We then introduce VBE-II, which only requires a single trajectory to perform the estimation, thus enjoying improved implementation simplicity.

\vspace{0.05in}
{\bf VBE-I.}
For a to-be-evaluated policy $\pi$, suppose we have collected trajectory $\xi_V$ for forming an estimator of $V^{\pi}$, denoted by 
$V^{\pi, \xi_V}$.
 We then generate another independent trajectory $\xi_Q$ of length $n+1$, 
which takes the form of $\cbr{Z_0, \ldots, Z_{n-1}, S_n} \equiv \cbr{(S_0, A_0), \ldots, (S_{n-1}, A_{n-1}), S_n }$.
Within trajectory $\xi_Q$, we define, for any $s \in \cS$, 
$\cT_n(s) \coloneqq  \cbr{t \leq n-1: S_t = s} $,
and 
\begin{align}\label{hitting_s_time_vbe}
\tau (s) \coloneqq \begin{cases}
 \min \cT_n(s), & ~ \cT_n(s) \neq \emptyset, \\
 n, & ~ \cT_n(s)  = \emptyset.
 \end{cases}
\end{align}

By denoting $\xi = (\xi_V, \xi_Q)$, we now introduce the estimator constructed by VBE-I operator, defined as 
\begin{align}\label{q_vbe_raw_form}
Q^{\pi, \xi, n}_{\mathrm{VBE-I}} (s,a)  \coloneqq 
\begin{cases}
\frac{\mathbbm{1}_{\{ a = A_{\tau(s)}\}}}{\pi(a|s)} 
\sbr{ c(s,a) + \gamma V^{\pi, \xi_V}(S_{\tau(s) + 1}) }, & \tau(s) \leq n - 1, \\
0, & \tau(s) = n.
\end{cases}
\end{align}
As the above estimator does not use $(A_n, S_{n+1})$ in its construction, 
for simplicity of presentation, we re-define $A_{n} \coloneqq \tilde{a} \notin \cA$, and $S_{n+1} \coloneqq \tilde{s}$ for some fixed $\tilde{s} \in \cS$. This in turn implies  $\mathbbm{1}_{\{ a = A_n\}} = 0$ for any $a \in \cA$,
and hence $Q^{\pi, \xi, n}_{\mathrm{VBE-I}} (s,a)$ can be succinctly written as 
\begin{align*}
Q^{\pi, \xi, n}_{\mathrm{VBE-I}}(s,a) = \frac{\mathbbm{1}_{\{ a = A_{\tau(s)}\}}}{\pi(a|s)} 
\sbr{ c(s,a) + \gamma V^{\pi, \xi_V}(S_{\tau(s) + 1}) }.
\end{align*}

We now proceed to bound the bias of the above estimator.

\begin{lemma}\label{lemma_bias_vbe_generic}
For any  $s \in \cS$ with $\nu^\pi(s) > 0$, let $t_{\mathrm{mix}}(s) =  \ceil{\log_\rho ( \tfrac{\nu^\pi(s)}{2 C} )}$, then 
\begin{align*}
 \abs{
\EE_{\xi} Q^{\pi, \xi, n}_{\mathrm{VBE-I}}(s,a) - Q^{\pi}(s,a)
}
\leq 
 \norm{  \EE_{\xi_V}  V^{\pi, \xi_V}  - V^\pi}_\infty  +  { 4 \rbr{1 - {\nu^{\pi}(s)}/{2}}^{\ceil{n/t_{\mathrm{mix}}(s)}}}/\rbr{1-\gamma},
 ~\forall a \in \cA.
\end{align*}

\end{lemma}

\begin{proof}
By denoting $ \tilde{V}^{\pi}(\cdot) = \EE_{\xi_V}  V^{\pi, \xi_V}(\cdot)$, we proceed with the following decomposition of the bias:
\begin{align}
& \abs{
\EE_{\xi} Q^{\pi, \xi, n}_{\mathrm{VBE-I}}(s,a) - Q^{\pi}(s,a)
}  \nonumber \\
 \overset{(a)}{=} &
\abs{
\EE_{\xi_Q}  \frac{\mathbbm{1}_{\{ a = A_{\tau(s)}\}}}{\pi(a|s)} 
\sbr{ c(s,a) + \gamma \tilde{V}^{\pi} (S_{\tau(s) + 1}) }
- \rbr{
c(s,a) + \gamma \EE_{s' \sim \cP(\cdot|s,a)} V^{\pi}(s') 
}
} \nonumber \\
\leq &
\abs{
\EE_{\xi_Q}  \frac{\mathbbm{1}_{\{ a = A_{\tau(s)}\}}}{\pi(a|s)}  - 1 
}
+ 
\gamma \abs{
\EE_{\xi_Q}  \frac{\mathbbm{1}_{\{ a = A_{\tau(s)}\}}}{\pi(a|s)} \tilde{V}^{\pi} (S_{\tau(s) + 1})
- \EE_{s' \sim \cP(\cdot|s,a)} V^{\pi}(s') 
} \label{value_vased_q_bias_decomp},
\end{align}
where $(a)$ uses the independence of $\xi_V$ and $\xi_Q$.
For the first term above, 
we have 
\begin{align}
& \abs{
\EE_{\xi_Q}  \frac{\mathbbm{1}_{\{ a = A_{\tau(s)}\}}}{\pi(a|s)}  - 1 
}  \nonumber \\ 
 \leq&  \tsum_{t=0}^{n-1} \PP(\tau(s) = t) \big|
\EE_{\xi_Q} \big[
 \frac{\mathbbm{1}_{\{ a = A_{\tau(s)}\}}}{\pi(a|s)} - 1 \big| \tau(s) = t
\big]
 \big|  
 + 
 \PP(\tau(s) = n) \big|
\EE_{\xi_Q} \big[
 \frac{\mathbbm{1}_{\{ a = A_{\tau(s)}\}}}{\pi(a|s)} - 1 \big| \tau(s) = n
\big]
 \big|
 \nonumber \\ 
\overset{(b)}{\leq} & \PP(\tau(s) = n)
 \overset{(c)}{\leq} 2  \rbr{1 - \frac{\nu^{\pi}(s)}{2}}^{\ceil{n/t_{\mathrm{mix}}(s)} }, \label{value_vased_q_bias_term1}
\end{align}
where 
$(b)$ follows from  $A_{\tau(s)} \sim \pi(\cdot|s)$ conditioned on $\tau(s) = t \leq n-1$, 
and the fact that $\mathbbm{1}_{\cbr{a = A_{\tau(s)}}} = 0$ conditioned on $\tau(s) = n$, given the definition of $A_n$;  
$(c)$ follows from the exact same lines as in \eqref{prob_for_hitting_time_z}, with 
$t_{\mathrm{mix}}=  \ceil{\log_\rho ( \tfrac{\nu^\pi(s)}{2 C})}$.
Similarly, we can bound the second term in \eqref{value_vased_q_bias_decomp} as
\begin{align}
& \abs{
\EE_{\xi_Q}  \frac{\mathbbm{1}_{\{ a = A_{\tau(s)}\}}}{\pi(a|s)} \tilde{V}^{\pi} (S_{\tau(s) + 1})
- \EE_{s' \sim \cP(\cdot|s,a)} V^{\pi}(s') 
} \nonumber \\
\leq & 
\tsum_{t=0}^{n-1} \PP(\tau(s) = t) \abs{
\EE_{\xi_Q}  \sbr{  \frac{\mathbbm{1}_{\{ a = A_{\tau(s)}\}}}{\pi(a|s)} \tilde{V}^{\pi} (S_{\tau(s) + 1})
- \EE_{s' \sim \cP(\cdot|s,a)} V^{\pi}(s') \big| 
\tau(s) = t
}
}
\nonumber \\ 
& ~~~~~~ + \PP(\tau(s) = n) \abs{
\EE_{\xi_Q}  \sbr{  \frac{\mathbbm{1}_{\{ a = A_{\tau(s)}\}}}{\pi(a|s)} \tilde{V}^{\pi} (S_{\tau(s) + 1})
- \EE_{s' \sim \cP(\cdot|s,a)} V^{\pi}(s') \big| 
\tau(s) = n
}
}
\nonumber \\
 \overset{(d)}{\leq} &
 \PP(\tau(s) = n) \big|
\EE_{\xi_Q}  \sbr{  \frac{\mathbbm{1}_{\{ a = A_{\tau(s)}\}}}{\pi(a|s)} \tilde{V}^{\pi} (S_{\tau(s) + 1}) - \EE_{s' \sim \cP(\cdot|s,a)} V^{\pi}(s') \big| 
\tau(s) = n
}
\big|
\nonumber \\ & ~~~ + \PP(\tau(s) \leq n-1) \norm{\tilde{V}^{\pi} - V^\pi}_\infty   \nonumber \\
\overset{(e)}{\leq} &
\PP(\tau(s) \leq n-1) \norm{\tilde{V}^{\pi} - V^\pi}_\infty 
+ \frac{\PP(\tau(s) = n) }{1-\gamma} \nonumber \\
 \leq &
 \norm{\tilde{V}^{\pi} - V^\pi}_\infty  +  \frac{2 \rbr{1 - {\nu^{\pi}(s)}/{2}}^{\ceil{n/t_{\mathrm{mix}}(s)}}}{1-\gamma}, \label{value_vased_q_bias_term2}
\end{align} 
where 
$(d)$ follows from the fact that by conditioning on $\tau(s) = t \leq n-1$, and invoking the strong Markov property, we have
\begin{align*}
\EE_{\xi_Q} \sbr{
 \frac{\mathbbm{1}_{\{ a = A_{\tau(s)}\}}}{\pi(a|s)} \tilde{V}^{\pi} (S_{\tau(s) + 1}) \big| \tau(s) = t
}
= \EE_{A \sim \pi(\cdot|s) } \EE_{S \sim \cP(\cdot|s,a) } \sbr{  \frac{\mathbbm{1}_{\{ a = A \}}}{\pi(a|s)} \tilde{V}^{\pi} (S)   }
= \EE_{s' \sim \cP(\cdot|s,a)} \sbr{\tilde{V}^{\pi}(s')}.
\end{align*}
In addition,  $(e)$ follows from  $\mathbbm{1}_{\{a = A_{\tau(s)\}}} = 0$ conditioned on $\tau(s) = n$, together with $0 \leq {V}^{\pi}(\cdot) \leq \frac{1}{1-\gamma}$.
Combining \eqref{value_vased_q_bias_term1}, \eqref{value_vased_q_bias_term2} with \eqref{value_vased_q_bias_decomp},  the desired claim follows immediately.
\end{proof}

In view of Lemma \ref{lemma_bias_vbe_generic}, the bias of the VBE estimator $Q^{\pi, \xi}$ decays linearly with respect to the number of samples, provided the bias of the value estimator $V^{\pi, \xi_V}$ does so.
In particular, this holds when $V^{\pi, \xi_V}$ is constructed by the on-policy Monte Carlo method defined in \eqref{OMC_for_value}.

\begin{corollary}\label{corr_q_vbe_bias}
Suppose Assumption \ref{assump_unif_mixing} holds. 
Let 
\begin{align}\label{eq_vbe_q_with_vbe_v}
Q^{\pi, \xi, n}_{\mathrm{VBE-I}} (s,a)  \coloneqq 
\begin{cases}
\frac{\mathbbm{1}_{\{ a = A_{\tau(s)}\}}}{\pi(a|s)} 
\sbr{ c(s,a) + \gamma V^{\pi, \xi_V, (n,1)}_{\mathrm{OMC}}(S_{\tau(s) + 1}) }, & \tau(s) \leq n - 1, \\
0, & \tau(s) = n,
\end{cases}
\end{align}
where $V^{\pi, \xi_V, (n,1)}_{\mathrm{OMC}}$ is defined in \eqref{OMC_for_value}. 
In addition, let $t_{\mathrm{mix}} = \ceil{\log_\rho ( \tfrac{\underline{\nu}}{2 C})}$.
Then we have  
\begin{align*}
 \norm{
\EE_{\xi} Q^{\pi, \xi, n}_{\mathrm{VBE-I}} - Q^{\pi} 
}_\infty \leq 
 \tfrac{2(n+1)}{1-\gamma} \sbr{ \gamma^{n-1} + (1- \tfrac{\underline{\nu}}{2 t_{\mathrm{mix}}})^{n-1}}  +   \frac{ 4 \rbr{1 - \underline{\nu}/{2}}^{\ceil{n/t_{\mathrm{mix}}}}}{1-\gamma}.
\end{align*}
\end{corollary}

\begin{proof}
The desired claim follows from a direct application of Lemma \ref{lemma_omc_value_est_bias} and Lemma \ref{lemma_bias_vbe_generic}.
\end{proof}

\begin{remark}
It is worth noting that 
the two-step construction of VBE-I also makes it feasible  to utilize other online policy evaluation operator for constructing $V^{\pi, \xi_V}$ with linearly converging bias.
This is particularly helpful as it can allow the policy evaluation step to incorporate simple  function approximation  for problems with large state spaces (e.g., linear approximation, \cite{li2022stochastic, kotsalis2022simple}).
This also seems to be a unique feature of VBE-I operator compared to other policy evaluation operators studied in this manuscript.
\end{remark}

\vspace{0.05in}
{\bf VBE-II.}
Compared to VBE-I, VBE-II only requires maintaining a single trajectory,
thus enjoying improved implementation simplicity.
The feasibility of this simplification stems from the observation that the estimated value in \eqref{eq_vbe_q_with_vbe_v} can be replaced by the future discounted cost starting from state $S_{\tau(s) + 1}$. 
In particular, 
for a trajectory $\xi$ of length $n+1$, 
which takes the form of $\cbr{Z_0, \ldots, Z_{n-1}, Z_n}$, 
the  VBE-II operator constructs the stochastic estimate as 
\begin{align}\label{q_vbe_type_ii}
Q^{\pi, \xi, n}_{\mathrm{VBE-II}} (s,a)  \coloneqq 
\begin{cases}
\frac{\mathbbm{1}_{\{ a = A_{\tau(s)}\}}}{\pi(a|s)} 
\cdot { \tsum_{t= \tau(s)}^{n-1} \gamma^{t - \tau(s)} c(S_t, A_t) }, & \tau(s) \leq n - 1, \\
0, & \tau(s) = n,
\end{cases}
\end{align}
where $\tau(s)$ is defined as in \eqref{hitting_s_time_vbe}. 
We now proceed to establish the bias of the VBE-II estimator.

\begin{lemma}\label{lemma_vbe_type_ii_bias}
Suppose Assumption \ref{assump_unif_mixing} holds. 
Then the VBE-II estimator \eqref{q_vbe_type_ii} satisfies 
\begin{align*}
\norm{
\EE_{\xi} Q^{\pi, \xi, n}_{\mathrm{VBE\text{-}II}} - Q^{\pi} 
}_\infty \leq 
 \tfrac{2(n+1)}{1-\gamma} \sbr{ \gamma^{n-1} + (1- \tfrac{\underline{\nu}}{2 t_{\mathrm{mix}}})^{n-1}} ,
\end{align*}
where $t_{\mathrm{mix}} = \ceil{\log_\rho ( \tfrac{\underline{\nu}}{2 C})}$.
\end{lemma}

\begin{proof}
Let us denote $\xi_{: \tau(s) } = \cbr{Z_{0}, \ldots, Z_{\tau(s)} }$,  $\xi_{\tau(s): } = \cbr{Z_{\tau(s) + 1}, \ldots, Z_{n} }$.
Since $\tau(s)$ is a stopping time, from the strong Markov property it follows that, 
\begin{align}\label{vbe_type_ii_conditional_bias}
\abs{
\EE_{\xi_{\tau(s):} | \tau(s), A_{\tau(s)} } \sbr{ \tsum_{t= \tau(s)}^{n-1} \gamma^{t - \tau(s)} c(S_t, A_t)
} - Q^{\pi}(s, A_{\tau(s)}) 
} \cdot \mathbbm{1}_{\{ \tau(s) \leq n-1\}}
\leq \frac{1}{1-\gamma} \gamma^{n - \tau(s)} .
\end{align}
Given the above relation,   we obtain
\begin{align*}
& \abs{
\EE\sbr{Q^{\pi, \xi, n}_{\mathrm{VBE-II}} (s,a)
} - Q^{\pi}(s, a) 
} \\
= & \abs{
\EE_{\tau(s)} \mathbbm{1}_{\{\tau(s) \leq n-1\}}  \EE_{A_{\tau(s)} | \tau(s)} \sbr{
 \frac{\mathbbm{1}_{\{ a = A_{\tau(s)}\}}}{\pi(a|s)}  \cdot \EE_{\xi_{\tau(s):  } | \tau(s), A_{\tau(s) }}
\sbr{
 { \tsum_{t= \tau(s)}^{n-1} \gamma^{t - \tau(s)} c(S_t, A_t) }
 }
 } 
 - Q^{\pi}(s,a)
} \\
\overset{(a)}{=}  & \big|
\EE_{\tau(s)} \mathbbm{1}_{\{\tau(s) \leq n-1\}}  \EE_{A_{\tau(s)} | \tau(s)} \sbr{
 \frac{\mathbbm{1}_{\{ a = A_{\tau(s)}\}}}{\pi(a|s)}  \cdot \EE_{\xi_{\tau(s):  } | \tau(s), A_{\tau(s) }}
\sbr{
 { \tsum_{t= \tau(s)}^{n-1} \gamma^{t - \tau(s)} c(S_t, A_t) }
 }
 } \\
 & ~~~ - \EE_{\tau(s)} \mathbbm{1}_{\{\tau(s) \leq n-1\}}  \EE_{A_{\tau(s)} | \tau(s)}  \big[ \frac{\mathbbm{1}_{\{ a = A_{\tau(s)}\}}}{\pi(a|s)} Q^{\pi}(s, A_{\tau(s)})\big]
\big| \\
= & \abs{
\EE_{\tau(s)} \mathbbm{1}_{\{\tau(s) \leq n-1\}}  \EE_{A_{\tau(s)} | \tau(s)} \sbr{
 \frac{\mathbbm{1}_{\{ a = A_{\tau(s)}\}}}{\pi(a|s)}  \rbr{ \EE_{\xi_{\tau(s):  } | \tau(s), A_{\tau(s) }}
\sbr{
 { \tsum_{t= \tau(s)}^{n-1} \gamma^{t - \tau(s)} c(S_t, A_t) }
 }
  - Q^{\pi}(s, A_{\tau(s)})
 } 
 }
} \\
\overset{(b)}{\leq} & 
\EE_{\tau(s)}   \mathbbm{1}_{\{\tau(s) \leq n-1\}}  \cdot \EE_{A_{\tau(s)} | \tau(s)}
\sbr{
\frac{\mathbbm{1}_{\{ a = A_{\tau(s)}\}}}{\pi(a|s)} \cdot 
\frac{1}{1-\gamma} \gamma^{n - \tau(s)} 
} \\
\overset{(c)}{=} & 
\EE_{\tau(s)}  \sbr{ \frac{1}{1-\gamma} \gamma^{n - \tau(s)} } \\
\overset{(d)}{\leq} & 
\tfrac{2(n+1)}{1-\gamma} \sbr{ \gamma^{n-1} + (1- \tfrac{\underline{\nu}}{2 t_{\mathrm{mix}}})^{n-1}},
\end{align*}
where 
$(a)$ follows from that $A_{\tau(s)} \sim \pi(\cdot|s)$ conditioned on $\tau(s) \leq n-1$,
 in view of the definition of $\tau(s)$ in \eqref{hitting_s_time_vbe}, 
 from which we obtain
\begin{align*}
\mathbbm{1}_{\{\tau(s) \leq n-1\}} \cdot  \EE_{A_{\tau(s)} | \tau(s)}  \big[ \frac{\mathbbm{1}_{\{ a = A_{\tau(s)}\}}}{\pi(a|s)} Q^{\pi}(s, A_{\tau(s)}) \big] = Q^{\pi}(s,a); 
\end{align*}
In addition, $(b)$ follows from \eqref{vbe_type_ii_conditional_bias}, and
$(c)$ follows from $ \mathbbm{1}_{\{\tau(s) \leq n-1\}}  \cdot \EE_{A_{\tau(s)}|\tau(s)} [{{\mathbbm{1}_{\{ a = A_{\tau(s)}\}}}/{\pi(a|s)}}] = 1$.
Finally, $(d)$ follows from the same lines as in Lemma \ref{lemma_bias_omc},
with  $t_{\mathrm{mix}} = \ceil{\log_\rho ( \tfrac{\underline{\nu}}{2 C})}$ replacing $t_{\mathrm{mix}} (z)$ therein.
\end{proof}

Going forward, we reserve the term VBE,  for either the VBE-I defined in  \eqref{eq_vbe_q_with_vbe_v}, or VBE-II defined in \eqref{q_vbe_type_ii}.
We now turn our attention to establish some generic convergence properties of SPMD with VBE operator. 
Note that given Corollary \ref{corr_q_vbe_bias} and Lemma \ref{lemma_vbe_type_ii_bias}, 
both types of VBE estimators share the same upper bound on their biases. 
This allows us to provide a unified treatment of both estimators when employed by the SPMD method as the policy evaluation subroutine.

\subsection{Global Convergence of SPMD with VBE}
We now proceed to establish the global convergence of SPMD with the VBE operator.
It should be noted that our development in this section is tailored to the Kullback-Leibler divergence.

\begin{definition}[Kullback-Leibler divergence]
Let the distance-generating function in  \eqref{def:kl_bregman} be chosen as $w(\pi(\cdot|s)) = \tsum_{a \in \cA} \pi(a|s) \log \pi(a|s)$ for $\pi(\cdot|s) \in \mathrm{ReInt}(\Delta_{\cA})$,
and we extend the domain of $w(\cdot)$ to $\Delta_{\cA}$ by defining $0 = 0 \log 0$.
Then 
\begin{align}\label{eq_def_kl}
D^{\pi}_{\pi'}(s) = \tsum_{a \in \cA} \pi(a|s) \log (\tfrac{\pi(a|s)}{\pi'(a|s)}) , ~ \forall \pi, \pi' \in \Pi,
\end{align}  
which  corresponds the Kullback–Leibler (KL) divergence between  $\pi(\cdot|s)$ and $\pi'(\cdot|s)$.
\end{definition}

It might be worth mentioning that the update \eqref{spmd_update} of SPMD with $D^{\pi}_{\pi'}$ being the KL divergence admits a closed-form expression:
\begin{align}\label{close_form_with_kl}
\pi_{k+1}(a|s) \propto \pi_k(a|s) \exp \rbr{ - \eta_k Q^{\pi_k, \xi_k}(s, a)}, ~\forall (s,a) \in \cS \times \cA,
\end{align}
which  coincides with the update of the stochastic variant of natural policy gradient method \cite{agarwal2020optimality}.
The following performance difference lemma \cite{lan2022policy, kakade2002approximately} will prove useful for the remainder of our technical development.

\begin{lemma}\label{lemma:performance_diff}
For any pair of policies $\pi, \pi'$, 
\begin{align*}
V^{\pi'}(s) - V^{\pi}(s) = \tfrac{1}{1-\gamma} \EE_{s' \sim d_s^{\pi'}} \sbr{\inner{Q^{\pi}( s', \cdot) }{ \pi'(\cdot|s') - \pi(\cdot| s') }  }.
\end{align*}
\end{lemma}

Our ensuing discussions also make use of the following simple technical observation.
\begin{lemma}\label{lemma_weighted_kl_control}
For any $x, y \in \mathrm{ReInt}(\Delta_{\cA})$, and any $g \in \RR^{\abs{\cA}}_{+}$, we have 
\begin{align*}
\inner{g}{x-y} - \tsum_{i \in \cA} y_i \log \rbr{ {y_i}/{x_i} }
\leq \tsum_{i \in \cA} {x_i g_i^2}/{2}.
\end{align*}
\end{lemma}

\begin{proof}
Consider the problem  
\begin{align}\label{kl_lemma_weighted_raw_formulation}
\textstyle
 \max_{y \in  \RR^{\abs{\cA}}}   \cL(y) \coloneqq  \inner{g}{x-y} - \tsum_{i \in \cA} y_i \log \rbr{ {y_i}/{x_i} },
~\mathrm{s.t.}~ \sum_{i \in \cA} y_i = 1, ~ y_i \geq 0, ~\forall i \in \cA,
\end{align}
which is concave, upper bounded,  and satisfies the Slater condition. By strong duality, there exists $\lambda \in \RR$, such that the solution $y^*$ of \eqref{kl_lemma_weighted_raw_formulation} satisfies 
\begin{align*}
\textstyle
y^* =  \argmax_{y \in \RR_+^{\abs{\cA}}}   \inner{g}{x-y} - \tsum_{i \in \cA} y_i \log \rbr{ {y_i}/{x_i} } + \lambda \rbr{\tsum_{i \in \cA} y_i - 1}.
\end{align*}
The first-order optimality condition of the above problem yields 
$
\log y_i^* = \log x_i - g_i - 1 + \lambda$, for all $ i \in \cA$.
From this we obtain $1 \equiv \sum_{i \in \cA} y_i^* = \exp(\lambda -1) \sum_{i \in \cA} x_i \exp(-g_i)$, and hence 
$1 - \lambda  = \log \rbr{\sum_{i \in \cA} x_i \exp(-g_i)}$.  
Substituting  the above characterizations of $y_i^*$ and $\lambda$ into \eqref{kl_lemma_weighted_raw_formulation}, we obtain
\begin{align*}
\cL(y^*) = (1-\lambda) + \inner{g}{x} 
& =  \log \rbr{\tsum_{i \in \cA} x_i \exp(-g_i)} + \inner{g}{x} \\
& \overset{(a)}{\leq} \tsum_{i \in \cA} x_i \exp(-g_i)  + \inner{g}{x} - 1 \\
& \overset{(b)}{\leq} \tsum_{i \in \cA} x_i \rbr{1 - g_i + g_i^2/2}  + \inner{g}{x} - 1 \\
& \overset{(c)}{=} \tsum_{i \in \cA} x_i g_i^2/2,
\end{align*} 
where $(a)$ follows from $\log(x) \leq x -1$ for $x > 0$, 
$(b)$ follows from $\exp(-x) \leq 1 - x + x^2/2$ for $x \geq 0$, and $(c)$ follows from $x \in \Delta_{\cA}$.
The proof is then completed.
\end{proof}

With Lemma \ref{lemma_weighted_kl_control} in place, the following lemma characterizes each update of SPMD with VBE operator.

\begin{lemma}\label{lemma:three_point_vbe}
For any $p \in \Pi$, the policy pair $(\pi_k, \pi_{k+1})$ in SPMD satisfies
\begin{align}\label{ineq:three_point_vbe}
\eta_k \big[ \inner{Q^{\pi_k, \xi_k}(s, \cdot)}{ \pi_{k+1} (\cdot| s )- p (\cdot| s)} \big]  + D^{\pi_{k+1}}_{\pi_k} (s) 
  \leq D^p_{\pi_k} (s) -  D^p_{\pi_{k+1}} (s) ,  ~ \forall s \in \cS.
\end{align}
In addition, with $D^{\pi_{k+1}}_{\pi_k}(s)$ being the KL divergence defined in \eqref{eq_def_kl}, we have 
\begin{align}\label{ineq_nonsmooth_vi_vbe}
& \eta_k \inner{Q^{\pi_k}(s, \cdot)}{\pi_k(\cdot|s) - \pi^*(\cdot|s)}  \nonumber \\
\leq &  D^{\pi^*}_{\pi_k} (s) -  D^{\pi^*}_{\pi_{k+1}} (s)
+  { \eta_k^2 \tsum_{a \in \cA} \pi_k(a|s) Q^{\pi_k, \xi_k}(s, a)^2} / {2} 
+ \eta_k \inner{\delta_k(s, \cdot)}{\pi_k(\cdot |s) - \pi^*(\cdot|s)} ,
\end{align}
where $\delta_k \coloneqq Q^{\pi_k} -  Q^{\pi_k, \xi_k}$.
\end{lemma}

\begin{proof}
From the optimality condition of the update \eqref{spmd_update}, it holds that 
\begin{align*}
 \inner{\eta_k Q^{\pi_k, \xi_k}(s, \cdot) + \partial D^{\pi_{k+1}}_{\pi_k} (s) }{\pi_{k+1}(\cdot|s) - p(\cdot|s) } \leq 0,
\end{align*}
where $ \partial D^{\pi_{k+1}}_{\pi_k} (s)$ denotes the subgradient of $ D^{\pi_{k+1}}_{\pi_k}(s)$ w.r.t. $\pi_{k+1}(\cdot|s)$. 
Then \eqref{ineq:three_point} follows by noting that
$
 \inner{\partial D^{\pi_{k+1}}_{\pi_k} (s) }{\pi_{k+1}(\cdot|s) - p(\cdot|s) } = -D^p_{\pi_k} (s) + D^p_{\pi_{k+1}} (s) + D^{\pi_{k+1}}_{\pi_k} (s) .
$
In addition, taking $p = \pi^*$ in \eqref{ineq:three_point_vbe} yields
\begin{align*}
& \eta_k \inner{Q^{\pi_k}(s, \cdot)}{\pi_k(\cdot|s) - \pi^*(\cdot|s)}  \nonumber \\
\leq &  D^{\pi^*}_{\pi_k} (s) -  D^{\pi^*}_{\pi_{k+1}} (s)
+ \eta_k \inner{Q^{\pi_k, \xi_k}}{\pi_k(\cdot|s) - \pi_{k+1}(\cdot|s)}
- D^{\pi_{k+1}}_{\pi_k}(s) 
+ \eta_k \inner{\delta_k(s, \cdot)}{\pi_k(\cdot |s) - \pi^*(\cdot|s)} .
\end{align*} 
From the fact that $c(s,a) \geq 0$ for any $(s,a) \in \cS \times \cA$, it holds $Q^{\pi}(s,a) \geq 0$ for any $\pi \in \Pi$.
 Then \eqref{ineq_nonsmooth_vi_vbe}  follows from 
applying Lemma \ref{lemma_weighted_kl_control} to the above relation.
\end{proof}

We are ready to establish the global convergence of SPMD with VBE as the policy evaluation subroutine. 

\begin{theorem}\label{thrm_spmd_vbe}
Consider SPMD with the VBE operator, defined either in \eqref{eq_vbe_q_with_vbe_v} or \eqref{q_vbe_type_ii}, as the policy evaluation subroutine.
Let the Bregman divergence be chosen as the KL divergence defined in \eqref{eq_def_kl}. 
For any $k > 0$, set $\eta_t = \eta \coloneqq  \sqrt{\frac{2 (1-\gamma)^2 \log \abs{\cA}}{k \abs{\cA}}}$ for all $t \leq k-1$, then we have 
\begin{align}\label{spmd_vbe_opt_gap}
\EE \sbr{ f({\pi}_R) - f(\pi^*) }
\leq  \sqrt{ \frac{2 \abs{\cA} \log \abs{\cA}}{k (1-\gamma)^4}}
+ \frac{\varepsilon_n }{1-\gamma} ,
\end{align}
where $\varepsilon_n = \tfrac{4(n+1)}{1-\gamma} \sbr{ \gamma^{n-1} + (1- \tfrac{\underline{\nu}}{2 t_{\mathrm{mix}}})^{n-1}}  +   \frac{ 8 \rbr{1 - \underline{\nu}/{2}}^{\ceil{n/t_{\mathrm{mix}}}}}{1-\gamma}$,
and $R \sim \mathrm{Unif}(\cbr{0, \ldots, k-1})$.
In addition, to obtain $\EE \sbr{ f({\pi}_R) - f(\pi^*) } \leq \epsilon$, the total number of samples can be bounded by 
\begin{align}\label{spmd_vbe_samples}
\tilde{\Theta} \rbr{
 \frac{\abs{\cA} \log \abs{\cA}}{\epsilon^2 (1-\gamma)^4}
 \rbr{ \frac{1}{1-\gamma} +  \frac{t_{\mathrm{mix}}}{\underline{\nu}} } \log \rbr{\frac{1}{(1-\gamma) \epsilon }} 
}.
\end{align}
\end{theorem}

\begin{proof}
Taking expectation with respect to $s \sim d_{\vartheta}^{\pi^*}$ in \eqref{ineq_nonsmooth_vi_vbe} of Lemma \ref{lemma:three_point_vbe}, we obtain 
\begin{align*}
& \eta_t (1-\gamma) \rbr{f(\pi_t) - f(\pi^*)} \\
  \overset{(a)}{=} & \eta_t \EE_{s \sim \vartheta} \EE_{s' \sim d_{s}^{\pi^*}} \inner{Q^{\pi_t}(s', \cdot)}{\pi_t(\cdot|s') - \pi^*(\cdot|s')}   \\
\overset{(b)}{ \leq} &    \EE_{s \sim d_{\vartheta}^{\pi^*}}  D^{\pi^*}_{\pi_t} (s) -    \EE_{s \sim d_{\vartheta}^{\pi^*}}  D^{\pi^*}_{\pi_{t+1}} (s)
+  \eta_t \EE_{s \sim d_{\vartheta}^{\pi^*}}   \inner{\delta_t(s, \cdot)}{\pi_t(\cdot |s) - \pi^*(\cdot|s)}  \\
& ~~~ +   \eta_t^2 \EE_{s \sim d_{\vartheta}^{\pi^*}} \tsum_{a \in \cA} \pi_t(a|s) Q^{\pi_t, \xi_t}(s, a)^2 / 2 .
\end{align*} 
where $(a)$ uses Lemma \ref{lemma:performance_diff}, 
$(b)$ uses  $d_{\vartheta}^{\pi^*}(s') = \EE_{s \sim \vartheta} d_s^{\pi^*}(s')$.
Further taking expectation with respect to $\xi_t$ in the above relation yields 
\begin{align}
& \eta_t (1-\gamma) \rbr{f(\pi_t) - f(\pi^*)}  \nonumber \\
\leq &    \EE_{s \sim d_{\vartheta}^{\pi^*}}  D^{\pi^*}_{\pi_t} (s) -    \EE_{s \sim d_{\vartheta}^{\pi^*}}  \EE_{\xi_t}  D^{\pi^*}_{\pi_{t+1}} (s)
+ \eta_t   \EE_{s \sim d_{\vartheta}^{\pi^*}} \EE_{\xi_t}  \inner{\delta_t(s, \cdot)}{\pi_t(\cdot |s) - \pi^*(\cdot|s)} \nonumber \\
& ~~~ +   \eta_t^2  \EE_{s \sim d_{\vartheta}^{\pi^*}} \EE_{\xi_t} \tsum_{a \in \cA} \pi_t(a|s) Q^{\pi_t, \xi_t}(s, a)^2 / 2 . 
\label{ineq_vbr_opt_gap_raw}
\end{align}
Given Corollary \ref{corr_q_vbe_bias} and Lemma \ref{lemma_vbe_type_ii_bias},  the third term in \eqref{ineq_vbr_opt_gap_raw} can be bounded by 
\begin{align}\label{ineq_vbr_control_bias}
&  \EE_{s \sim d_{\vartheta}^{\pi^*}} \EE_{\xi_t}  \inner{\delta_t(s, \cdot)}{\pi_t(\cdot |s) - \pi^*(\cdot|s)} \nonumber \\
  \leq & 
2   \norm{\EE_{\xi_t} \delta_t }_\infty 
\leq 
 \tfrac{4(n+1)}{1-\gamma} \sbr{ \gamma^{n-1} + (1- \tfrac{\underline{\nu}}{2 t_{\mathrm{mix}}})^{n-1}}  +   \frac{ 8 \rbr{1 - \underline{\nu}/{2}}^{\ceil{n/t_{\mathrm{mix}}}}}{1-\gamma}
 \equiv \varepsilon_n
 .
\end{align}
Turning our attention to the fourth term in \eqref{ineq_vbr_opt_gap_raw}, 
from \eqref{eq_vbe_q_with_vbe_v} and \eqref{q_vbe_type_ii}, it is clear that
for both VBE-I and VBE-II, 
\begin{align*}
\tsum_{a \in \cA} \pi_t(a|s) Q^{\pi_t, \xi_t}(s, a)^2  \leq 
\begin{cases}
\frac{1}{\pi_t(A_{\tau(s)} |s) (1-\gamma)^2} ,  &  \tau(s) \leq n-1, \\
0 , & \tau(s) = n.
\end{cases} 
\end{align*}
Hence by noting that $A_{\tau(s)} \sim \pi(\cdot|s)$ conditioned on $\tau(s) \leq n-1$, 
\begin{align}\label{ineq_vbe_control_wt_q}
&  \EE_{s \sim d_{\vartheta}^{\pi^*}} \EE_{\xi_t} \tsum_{a \in \cA} \pi_t(a|s) Q^{\pi_t, \xi_t}(s, a)^2   \nonumber \\
\leq &  \max_{s \in \cS} 
\EE_{\tau(s)} \EE_{A_{\tau(s)} | \tau(s) } \sbr{ \frac{1}{\pi_t(A_{\tau(s)} |s) (1-\gamma)^2}  \big| \tau(s) } \mathbbm{1}_{\{\tau(s) \leq n-1 \}}
= \tsum_{a \in \cA} \frac{\pi_t(a|s)}{\pi_t(a|s) (1-\gamma)^2} =
 {(1-\gamma)^{-2} \abs{\cA} }.
\end{align}
Combining \eqref{ineq_vbr_opt_gap_raw}, \eqref{ineq_vbr_control_bias}, and \eqref{ineq_vbe_control_wt_q}, we obtain 
\begin{align*}
& \eta_t (1-\gamma) \rbr{f(\pi_t) - f(\pi^*)}  \nonumber \\
\leq &    \EE_{s \sim d_{\vartheta}^{\pi^*}}  D^{\pi^*}_{\pi_t} (s) -    \EE_{s \sim d_{\vartheta}^{\pi^*}}  \EE_{\xi_t}  D^{\pi^*}_{\pi_{t+1}} (s)
+ \eta_t  \varepsilon_n +   \eta_t^2 {(1-\gamma)^{-2} \abs{\cA}} / {2} .
\end{align*}
Now taking total expectation of the above relation, and further taking the telescopic sum from $t =0$ to $k-1$ with constant stepsize $\eta_t = \eta > 0$,
it holds that 
\begin{align*}
\EE \sbr{ f({\pi}_R) - f(\pi^*) }
\leq \frac{\log \abs{\cA}}{\eta k (1-\gamma) } 
+ \frac{\varepsilon_n }{1-\gamma} 
+ \frac{\eta \abs{\cA}}{2 (1-\gamma)^3}.
\end{align*}
Setting $\eta = \sqrt{\frac{2 (1-\gamma)^2 \log \abs{\cA}}{k \abs{\cA}}}$ in the above relation
concludes the proof for \eqref{spmd_vbe_opt_gap}.
To obtain $\EE \sbr{ f({\pi}_R) - f(\pi^*) } \leq \epsilon$, 
if suffices to take 
$k = \frac{8 \abs{\cA} \log \abs{\cA}}{\epsilon^2 (1-\gamma)^4}$, and 
\begin{align*}
n = \tilde{\Theta} \rbr{
\rbr{ \frac{1}{1-\gamma} +  \frac{t_{\mathrm{mix}}}{\underline{\nu}} } \log \rbr{\frac{1}{(1-\gamma) \epsilon}} 
}
~ \Rightarrow 
~ 
\frac{\varepsilon_n}{1-\gamma} \leq \frac{\epsilon}{2}.
\end{align*}
The total number of samples is given by 
\begin{align*}
k \cdot n = 
\tilde{\Theta} \rbr{
 \frac{\abs{\cA} \log \abs{\cA}}{\epsilon^2 (1-\gamma)^4}
 \rbr{ \frac{1}{1-\gamma} +  \frac{t_{\mathrm{mix}}}{\underline{\nu}} } \log \rbr{\frac{1}{(1-\gamma) \epsilon }} 
},
\end{align*}
which concludes the proof for \eqref{spmd_vbe_samples}.
\end{proof}
In view of Theorem \ref{thrm_spmd_vbe}, the KL divergence-based
SPMD with the VBE operator attains an $\tilde{\cO}(1/\epsilon^2)$ sample complexity, which also exhibits a linear dependence on the size of the action space (up to a logarithmic factor).
The obtained sample complexity seems to be the first time that optimal dependence on the precision target is achieved among online PG methods with no explicit exploration strategies.
It might be worth noting that when $\abs{\cS} = 1$ and $\gamma = 0$, 
then KL divergence-based SPMD with the VBE operator recovers the EXP3 method introduced for the adversarial bandit problem \cite{auer2002nonstochastic}.

On the other hand,
it remains unclear whether  Theorem \ref{thrm_spmd_vbe} can be extended to general Bregman divergences. 
The obtained optimality gap \eqref{spmd_vbe_opt_gap} also only holds in expectation.
We believe the expectation bound of the optimality gap is intrinsic to the VBE operator, and can not be strengthened to a high probability bound without further modifications to the operator.
 Indeed, as the policy in SPMD approaches optimal policies, non-optimal actions will be associated with a diminishing policy value, and it can be readily verified that in this case both variants of the  VBE operator  \eqref{q_vbe_raw_form} and \eqref{eq_vbe_q_with_vbe_v} have unbounded variance. 

In the next section, we introduce the second policy evaluation operator that is compatible with a much larger class of Bregman divergences. 
In addition, one can further relax Assumption \ref{assump_unif_mixing} to a weaker condition.
More importantly, SPMD with this operator directly controls the optimality gap with high probability, as opposed to the expecation bound of the VBE operator. 

\section{SPMD with Truncated On-policy Monte Carlo}\label{sec_spmd_omc}


This section introduces the second online policy evaluation operator, named truncated on-policy Monte-Carlo (TOMC), for constructing the stochastic estimators $\{Q^{\pi_t, \xi_t}\}$ used in SPMD.
We will establish the global convergence of SPMD with the TOMC operator for a general class of Bregman divergences, and consequently determine its sample complexity.

\subsection{TOMC and a General Class of Bregman Divergences}

The TOMC operator outputs the following stochastic estimator $Q^{\pi_t, \xi_t}$, constructed by slightly modifying the OMC estimator of the Q-function defined in \eqref{omc_estimate_batch}, as
 \begin{align}\label{eq_truncate_q}
Q^{\pi_t, \xi_t} (s,a) = Q^{\pi_t, \xi_t, (n,m)}_{\mathrm{OMC}}(s,a) \cdot \mathbbm{1}_{\cbr{\pi_t(a|s) \geq \tau}} + \tfrac{1}{1-\gamma} \cdot \mathbbm{1}_{\cbr{\pi_t(a|s) < \tau}}.
\end{align}
That is, TOMC sets up a threshold $\tau > 0$, and for any action with its policy value falling below $\tau$, TOMC completely truncates the process of estimating $Q^{\pi_t}(s,a)$ using the collected samples, by simply discarding the samples and directly assigning an upper bound ${1}/\rbr{1-\gamma}$ as a trivial estimate.
The concrete specification of the defining parameters for TOMC, $(n, m, \tau)$, will be specified after we establish some generic convergence properties of SPMD with this evaluation operator.

\begin{remark}[Bias of TOMC Estimator]
The following difference between VBE and TOMC operators might be worthy of additional attention.
The VBE operator returns an almost unbiased estimate of the Q-value for every action, regardless of its policy value (cf. Corollary \ref{corr_q_vbe_bias} and Lemma \ref{lemma_vbe_type_ii_bias}).
In contrast,
in view of the truncation step in the construction of TOMC estimator  \eqref{eq_truncate_q}, 
 the bias of the TOMC estimator will be bounded away from zero for any action that has policy value below the threshold $\tau$.
 Notably, this non-trivial bias persists regardless the total number of samples collected for policy evaluation.
 As will be clear in our ensuing discussions, such an induced bias from the truncation step, contrary to being detrimental, is essential for SPMD to overcome the lack of exploration associated the vanilla OMC estimator (cf. Remark \ref{remark_vanilla_omc}).
\end{remark}

Unlike the VBE operator in Section \ref{sec_spmd_vbe}, TOMC operator enjoys much improved applicability, as it can be incorporated into SPMD with a  general class of Bregman divergences with the following properties.
\begin{condition}\label{divergence_condition_fixed_optimal}
The Bregman divergence \eqref{def:kl_bregman} satisfies  the following.
\begin{enumerate}
\item 
The distance-generating function $w$ is strongly-convex with respect to $\ell_1$-norm with modulus $\mu$:
\begin{align*}
D^{\pi}_{\pi'}(s) \geq  \tfrac{\mu}{2} \norm{\pi(\cdot|s) - \pi'(\cdot|s)}_1^2.
\end{align*}

\item For any $\pi_t \in \mathrm{ReInt}(\Pi)$, any stepsize $\eta_k \geq 0$,  and any $Q^{\pi_t, \xi_t} \in \RR^{\abs{\cS} \times \abs{\cA}}$, the SPMD update \eqref{spmd_update} always  yields 
$\pi_{t+1} \in \mathrm{ReInt}(\Pi)$.

\item
Fixing $\mathfrak{D} > 0$, 
if for a state $s \in \cS$,  a policy $\pi \in \mathrm{ReInt}(\Pi)$, and a deterministic optimal policy $\pi^* $,
 one has $(1-\gamma) D^{\pi^*}_{\pi} (s) \leq \mathfrak{D} $, 
 then there exists a constant $\tau(\mathfrak{D} ) > 0$ such that
\begin{align*}
\pi( a^{\pi^*}_s|s) \geq \tau(\mathfrak{D} ) ,
\end{align*}
where  $a^{\pi^*}_s$ denotes the unique action satisfying $\pi^*(a^{\pi^*}_s |s ) = 1$. 
\end{enumerate}
\end{condition}

The last part of Condition \ref{divergence_condition_fixed_optimal} states that if for a given state,  the policy $\pi$ has bounded Bregman divergence with an optimal deterministic policy $\pi^*$,
then the action taken by $\pi^*$ will also be taken by the policy $\pi$ with a probability bounded away from $0$.
 As we shall verify later, Condition \ref{divergence_condition_fixed_optimal} indeed can be satisfied by several practical Bregman divergences.
For the purpose of ensuing discussion, let us denote 
\begin{align*}
\textstyle
\overline{\Pi}^* = \cbr{\pi: \pi(a_s|s) = 1 ~\text{for some}~ a_s \in \cA^*_s, ~\forall s \in \cS}
\end{align*} 
 as the set of deterministic optimal policies, 
 where $\cA^*_s$ is defined in Lemma \ref{lemma:optimal_policy_set}.
Below, we establish some generic observations on the convergence properties of SPMD.

The next lemma provides a characterization on the policy update with general Bregman divergences.

\begin{lemma}\label{lemma:three_point}
For any $p \in \Pi$, the policy pair $(\pi_k, \pi_{k+1})$ in SPMD satisfies
\begin{align}\label{ineq:three_point}
\eta_k \big[ \inner{Q^{\pi_k, \xi_k}(s, \cdot)}{ \pi_{k+1} (\cdot| s )- p (\cdot| s)} \big]  + D^{\pi_{k+1}}_{\pi_k} (s) 
  \leq D^p_{\pi_k} (s) -  D^p_{\pi_{k+1}} (s) ,  ~ \forall s \in \cS.
\end{align}
In addition, with $\delta_k \coloneqq Q^{\pi_k} -  Q^{\pi_k, \xi_k}$, for any optimal policy $\pi^*$, we have
\begin{align}\label{ineq_nonsmooth_vi}
& \eta_k \inner{Q^{\pi_k}(s, \cdot)}{\pi_k(\cdot|s) - \pi^*(\cdot|s)}  \nonumber \\
\leq &  D^{\pi^*}_{\pi_k} (s) -  D^{\pi^*}_{\pi_{k+1}} (s)
+ \tfrac{\eta_k^2 \norm{Q^{\pi_k, \xi_k}}_\infty^2}{2 \mu} 
+ \eta_k \inner{\delta_k(s, \cdot)}{\pi_k(\cdot |s) - \pi^*(\cdot|s)} .
\end{align}
\end{lemma}

\begin{proof}
The first inequality \eqref{ineq:three_point} follows the exact same lines for showing \eqref{ineq:three_point_vbe}  in Lemma \ref{lemma:three_point_vbe}.
In addition, by taking $p= \pi^*$ in \eqref{ineq:three_point}, then \eqref{ineq_nonsmooth_vi} follows from 
\begin{align*}
& - D^{\pi_{k+1}}_{\pi_k} (s)  + \eta_k \inner{Q^{\pi_k, \xi_k}}{\pi_k(\cdot|s) - \pi_{k+1}(\cdot|s)}  \\
\leq & -\tfrac{\mu}{2} \norm{\pi_k(\cdot|s) - \pi_{k+1}(\cdot|s)}_1^2 + \eta_k  \norm{Q^{\pi_k, \xi_k}}_\infty \norm{\pi_k(\cdot|s) - \pi_{k+1}(\cdot|s)}_1
\leq \tfrac{\eta_k^2 \norm{Q^{\pi_k, \xi_k}}_\infty^2}{2 \mu} . 
\end{align*}
The proof is then completed.
\end{proof}

We proceed to establish an upper bound on the Bregman divergence between $\pi_k$ and any optimal policy $\pi^*$, provided the TOMC estimator \eqref{eq_truncate_q} satisfies  certain noise conditions.

\begin{lemma}\label{lemma_bounded_divergence}
For any $k > 0$, suppose the following holds for the stochastic estimator $Q^{\pi_t, \xi_t}$ output by the TOMC operator \eqref{eq_truncate_q} up to iterations $k-1$:
\begin{align}
\norm{Q^{\pi_t, \xi_t}}_\infty & \leq M, \label{condition_m_bound} \\
\inner{\delta_t(s, \cdot)}{\pi_t(\cdot |s) - \pi^*(\cdot|s)} & \leq \varepsilon,  ~\forall s \in \cS, \label{condition_bias_bound}
\end{align}
for some $M > 0$, $\varepsilon > 0$,  and $\pi^* \in \overline{\Pi}^*$. 
 Then with any constant stepsize $\eta > 0$, SPMD satisfies 
\begin{align*}
\textstyle
(1-\gamma) D^{\pi^*}_{\pi_{k}} (s)  \leq \max_{s' \in \cS} D^{\pi^*}_{\pi_0} (s')
+ \tfrac{\eta^2 M^2 k}{2 \mu} 
+  \eta k \varepsilon.
\end{align*}
\end{lemma}

\begin{proof}
By taking expectation of $s' \sim d_s^{\pi^*}$ in \eqref{ineq_nonsmooth_vi},
we obtain 
\begin{align}
& \eta_t ( 1-\gamma) \sbr{V^{\pi_t}(s) - V^{\pi^*}(s)} \nonumber  \\
 \overset{(a)}{=}  & 
\eta_t \EE_{s' \sim d_s^{\pi^*}}  \inner{Q^{\pi_t}(s', \cdot)}{\pi_t(\cdot|s') - \pi^*(\cdot|s')}  \nonumber \\
 \leq &  \EE_{s' \sim d_s^{\pi^*}} D^{\pi^*}_{\pi_t} (s') -  \EE_{s' \sim d_s^{\pi^*}} D^{\pi^*}_{\pi_{t+1}} (s')
+\tfrac{\eta_t^2 \norm{Q^{\pi_t, \xi_t}}_\infty^2}{2 \mu} 
+ \eta_t \EE_{s' \sim d_s^{\pi^*}}  \inner{\delta_t(s', \cdot)}{\pi_t(\cdot |s') - \pi^*(\cdot|s')},\label{divergence_to_optimal_bound}
\end{align}
where equality $(a)$ uses Lemma \ref{lemma:performance_diff}.
With constant stepsizes $\eta_t = \eta$, 
we sum up the above inequality from $0$ to $k-1$, which gives 
\begin{align*}
(1-\gamma) D^{\pi^*}_{\pi_{k}} (s) & \leq
\EE_{s' \sim d_s^{\pi^*}} D^{\pi^*}_{\pi_k} (s')  \leq \EE_{s' \sim d_s^{\pi^*}} D^{\pi^*}_{\pi_0} (s')
+ \tfrac{\eta^2 M^2 k }{2 \mu} 
+  \eta k \varepsilon,
\end{align*}
where the first inequality follows from $d_s^{\pi^*}(s) \geq 1-\gamma$,
and the second inequality follows from an immediate application of \eqref{condition_m_bound} and \eqref{condition_bias_bound}.
\end{proof}

The first condition \eqref{condition_m_bound} in Lemma \ref{lemma_bounded_divergence} is readily satisfied by the TOMC estimator with $M = \frac{1}{1-\gamma}$.
The second condition \eqref{condition_bias_bound}  can be met, for instance, if $\norm{\delta_t(s, \cdot)}_\infty$ is small, but requires the policy $\pi_t$ to explore every action. Consequently, this condition  breaks eventually given our discussion in Section \ref{spmd_technical_background}.
We will proceed with a more delicate approach that makes use of the truncation procedure in TOMC, and consequently remove this exploration requirement.
In particular, key parameters $(n, m ,\tau)$ of the TOMC estimator \eqref{eq_truncate_q} will be determined in a divergence dependent manner.

\subsection{SPMD with Multi-trajectory TOMC}\label{subsec_exp_multiple_traj_analysis}

This subsection 
starts our analysis by considering a relatively simple case, where one can use the TOMC operator \eqref{eq_truncate_q} with multiple independent trajectories ($m > 1$).
 We establish the global convergence of SPMD using this evaluation operator, and  obtain an $\tilde{\cO}(1/\epsilon^4)$ sample complexity for finding an $\epsilon$-optimal policy. 
 
 Though the sample complexity associated with multi-trajectory TOMC has a non-optimal dependence on the  target accuracy compared to that of the VBE operator, it is worth noting here that the optimality gap holds in high probability. 
 Indeed, this $\tilde{\cO}(1/\epsilon^4)$ sample complexity already improves upon existing online PG methods that can control the optimality gap in high probability without explicit exploration.
 TOMC is  also compatible to a general class of Bregman divergences satisfying Condition \ref{divergence_condition_fixed_optimal}. 
  Most importantly, discussions in this subsection gradually introduce several important conceptual ideas, which we build upon in Section \ref{subsec_single_traj}, leading to the optimal $\tilde{\cO}(1/\epsilon^2)$ sample complexity. 
  


We proceed with an inductive argument, which shows that 
for any $\varepsilon > 0$ and $\delta \in (0,1)$, 
with proper specification of $(n, m, \tau)$ of the TOMC estimator defined in \eqref{eq_truncate_q}, 
conditions \eqref{condition_m_bound} and \eqref{condition_bias_bound} in Lemma \ref{lemma_bounded_divergence} hold
up to iteration $t$, 
  with probability at least $(1-\delta)^{t+1}$, for any $t \geq 0$. 
As a consequence, we show that the accumulated noise in SPMD can be effectively bounded, an important step for establishing the global convergence of SPMD.

 \begin{lemma}\label{lemma_induction_bias_bound}
Suppose Assumption \ref{assump_unif_mixing} holds,
and the Bregman divergence \eqref{def:kl_bregman} satisfies Condition \ref{divergence_condition_fixed_optimal}.
  For any $\varepsilon > 0$,
let the total iterations $k > 0$ be given a priori, and set the stepsize $\eta_t =  \eta > 0$, with 
 \begin{align}\label{induction_stepsize_condition}
 \tfrac{\eta^2 M^2 t }{2 \mu}  \leq   \cD/4,
 ~
 \eta t \varepsilon \leq  \cD/4, ~ \text{for all $t \leq k$,}
\end{align}
where  $\cD > 0$ is any constant  satisfying  
$ \cD \geq 2 \max_{s' \in \cS} \max_{\pi^* \in \overline{\Pi}^*} D^{\pi^*}_{\pi_0} (s') $.
 Accordingly, let $\tau(\cD)$ be defined as in Condition \ref{divergence_condition_fixed_optimal}. 
For any $\delta \in (0,1)$,  
 let the parameters $(n, m, \tau)$ of the TOMC estimator \eqref{eq_truncate_q} be given as
\begin{align}\label{choice_m_n_general}
m = \Theta \rbr{\tfrac{\abs{\cA}^2}{(1-\gamma)^2 \varepsilon^2} \log (\tfrac{2 \abs{\cZ} }{\delta}) },
n = \tilde{\Theta} \rbr{
\rbr{ \tfrac{1}{1-\gamma} 
+\tfrac{\ceil{\log_\rho ( {\underline{\nu} \cdot {\tau}}/{(2 C)})}}{\underline{\nu}\cdot {\tau}}} \log (\tfrac{\abs{\cA} }{\varepsilon})
},
\tau = \min \cbr{\tau(\cD), 1/\abs{\cA}},
\end{align}
Then we have 
\begin{align*}
\PP(\cap_{i \leq t} \cE_i) \geq (1-\delta)^{t+1}, \text{ for all } t \leq k,
\end{align*}
 where $\cE_t \coloneqq \cbr{\omega: \sup_{s \in \cS} \inner{\delta_t(s, \cdot)}{\pi_t(\cdot |s) - \pi^*(\cdot|s)}  \leq \varepsilon}$. 
 \end{lemma} 
  
\begin{proof}

Let us fix an optimal deterministic policy $\pi^* \in \overline{\Pi}^*$,
and define $a_s^{\pi^*}$ as the unique action satisfying $\pi^*(a_s^{\pi^*} | s) =1$.
Consider the base case $t = 0$.
Given Assumption \ref{assump_unif_mixing} and the fact that $\pi_0$ is the uniform policy, we have  $\sigma^{\pi_0}(z) > \underline{\nu}/\abs{\cA} $ for all $z \in \cZ$. 
In addition, since $\tau \leq 1/\abs{\cA}$, we have $Q^{\pi_0, \xi_0} = Q^{\pi_0, \xi_0, (n, m)}_{\mathrm{OMC}}$ given \eqref{eq_truncate_q}.
Hence, in view of Lemma \ref{lemma_bias_omc} and the definition of $Q^{\pi_0, \xi_0, (n, m)}_{\mathrm{OMC}}$ in \eqref{omc_estimate_batch}, we invoke the Hoeffding's inequality together with a union bound over all $(s,a) \in \cZ$, which yields that for any $\delta > 0$, with probability $1-\delta$,  
\begin{align*}
\abs{\delta_0(s,a)} \leq \tfrac{2(n+1)}{1-\gamma} \sbr{ \gamma^{n-1} + (1- \tfrac{\sigma^\pi(z)}{2 t_{\mathrm{mix}(z)}})^{n-1}}
+ \tfrac{1}{1-\gamma} \sqrt{\tfrac{2 \log(2 \abs{\cZ}/\delta)}{2m } }, ~ \forall (s,a) \in \cZ.
\end{align*}
Noting that $t_{\mathrm{mix}}(z) = \ceil{\log_\rho ( \tfrac{\sigma^{\pi_0}(z)}{2 C})} \leq \ceil{\log_\rho ( \tfrac{\underline{\nu}  }{2 C \abs{\cA}})} $, we obtain 
\begin{align}\label{bias_from_hoeffding}
\abs{\delta_0(s,a)} \leq \tfrac{2(n+1)}{1-\gamma} \sbr{ \gamma^{n-1} + (1- \frac{\underline{\nu}}{2 \abs{\cA} \ceil{\log_\rho ( {\underline{\nu}  }/\rbr{2 C \abs{\cA}})}} )^{n-1}}
+ \tfrac{1}{1-\gamma} \sqrt{\tfrac{2 \log(2 \abs{\cZ}/\delta)}{2m } }.
\end{align}
Hence, by choosing 
\begin{align*}
m = \Theta \rbr{\tfrac{\abs{\cA}^2}{(1-\gamma)^2 \varepsilon^2} \log (\tfrac{2 \abs{\cZ} }{\delta}) },
n = \tilde{\Theta} \rbr{
\rbr{ \tfrac{1}{1-\gamma} 
+ \tfrac{\abs{\cA}  \ceil{\log_\rho ( {\underline{\nu}  }/\rbr{2 C \abs{\cA}})} }{\underline{\nu}} } \log (\tfrac{\abs{\cA}}{\varepsilon})
},
\end{align*}
we have \eqref{condition_bias_bound} holds with probability $1-\delta$.
Hence the induction hypothesis holds for $t=0$.

Now suppose the induction hypothesis holds up to iteration $t-1$. 
Then in view of Lemma \ref{lemma_bounded_divergence} and the induction hypothesis, 
with probability $(1-\delta)^t$, 
we have 
\begin{align*}
(1-\gamma) D^{\pi^*}_{\pi_{t}} (s)  \leq \max_{s' \in \cS} D^{\pi^*}_{\pi_0} (s')
+ \tfrac{\eta^2 M^2 t }{2 \mu} 
+  \eta t \varepsilon, ~\forall s \in \cS.
\end{align*}
Conditioned on $\cap_{i \leq t-1} \cE_i$, and
let 
$ \cD \geq 2 \max_{s' \in \cS, \pi^* \in \overline{\Pi}^*}   D^{\pi^*}_{\pi_0} (s') $, 
then given the choice of stepsize in \eqref{induction_stepsize_condition}, 
we  have $(1-\gamma) D^{\pi^*}_{\pi_{t}} (s)  \leq   \cD$.  
In view of Condition \ref{divergence_condition_fixed_optimal}, this in turn implies
\begin{align}\label{induction_prob_lb}
\pi_t(a^{\pi^*}_s |s) \geq \tau(\cD) > 0,  ~\forall s \in \cS.
\end{align}
By definition \eqref{choice_m_n_general}, it holds that $\tau \leq \tau(\cD)$.
Consequently, given \eqref{induction_prob_lb},  if $\pi_t (a|s) < \tau$,  it is immediate that $a \neq a^{\pi^*}_s$. 
Hence
\begin{align*}
& \inner{\delta_t(s, \cdot)}{\pi_t(\cdot |s) - \pi^*(\cdot|s)} \\
= & \tsum_{\pi_t(a|s) \geq \tau}   {\delta_t(s, a)}\rbr{\pi_t(a |s) - \pi^*(a|s)} 
+ \tsum_{\pi_t(a|s) < \tau} {\delta_t(s, a)}\rbr{\pi_t(a |s) - \pi^*(a|s)}  \\ 
\leq & 
\tsum_{\pi_t(a|s) \geq \tau}  2 \abs{\delta_t(s,a)}
+ \tsum_{\pi_t(a|s) < \tau} \delta_t(s,a)  \pi_t(a |s).
\end{align*}
From the definition
of $\delta_t$ in \eqref{def_delta_noise} and the TOMC estimator  \eqref{eq_truncate_q}, 
 it holds that $\delta_t(s,a) \leq 0$ if $\pi_t(a|s) < \tau$,  we can then drop the second term in the last inequality and obtain 
\begin{align*}
& \inner{\delta_t(s, \cdot)}{\pi_t(\cdot |s) - \pi^*(\cdot|s)}  \\
\leq & 
\tsum_{a \in \cA, \pi_t(a|s) \geq \tau}  2 \abs{\delta_t(s,a)} \\
\overset{(a)}{\leq} & \tfrac{4 \abs{\cA} (n+1)}{1-\gamma} \sbr{ \gamma^{n-1} + (1-  \frac{\underline{\nu} \tau}{ 2\ceil{\log_\rho ( {\underline{\nu} \tau}/\rbr{2 C})} })^{n-1}}
+ \tfrac{2\abs{\cA}}{1-\gamma} \sqrt{\tfrac{2 \log(2 \abs{\cZ}/\delta)}{2m } }, ~ \forall s \in \cS,
\end{align*}
with probability $1-\delta$,
where $(a)$ follows from the same arguments in \eqref{bias_from_hoeffding}.
Hence by 
choosing 
\begin{align*}
m = \Theta \rbr{\tfrac{\abs{\cA}^2}{(1-\gamma)^2 \varepsilon^2} \log (\tfrac{2 \abs{\cZ} }{\delta}) },
n = \tilde{\Theta} \rbr{
\rbr{ \tfrac{1}{1-\gamma} 
+ \tfrac{\ceil{\log_\rho ( {\underline{\nu} \cdot {\tau}}/{(2 C)})}}{\underline{\nu}\cdot {\tau}} } \log (\tfrac{\abs{\cA} }{\varepsilon})
},
\end{align*}
we have $\inner{\delta_t(s, \cdot)}{\pi_t(\cdot |s) - \pi^*(\cdot|s)} \leq \varepsilon$ for any $s \in \cS$, with probability $1-\delta$, conditioned on $\cap_{i \leq t-1} \cE_i$.
Combined with the induction hypothesis, this implies 
$\PP ( \cap_{i \leq t} \cE_i) \geq (1-\delta)^{t+1}$.
The  proof is then completed.
\end{proof}

\begin{remark}\label{remark_relax_assump}
It might be worth noting that, for the TOMC operator, Assumption \ref{assump_unif_mixing} can be relaxed in the following sense. 
Instead of requiring the conditions of Assumption \ref{assump_unif_mixing} to hold for all the policies $\cbr{\pi_t}$ generated by SPMD, it is sufficient to require the existence of a deterministic policy $\pi^* \in \overline{\Pi}^*$, such that Assumption \ref{assump_unif_mixing} holds for this single policy $\pi^*$.
Indeed, in view of observation \eqref{induction_prob_lb}, 
every action taken by  $\pi^*$ will be taken by the generated policy $\pi_t$ with a probability bounded away from 0, consequently 
 it is not difficult to show that this relaxed condition implies Assumption \ref{assump_unif_mixing}.
\end{remark}

With Lemma \ref{lemma_induction_bias_bound} in place, we are now ready to establish the global convergence of SPMD with the TOMC operator that uses multiple trajectories.

\begin{theorem}\label{global_convergence_generic}
Suppose Assumption \ref{assump_unif_mixing} holds,
and the Bregman divergence \eqref{def:kl_bregman} satisfies Condition \ref{divergence_condition_fixed_optimal}.
Fix total iterations $k > 0$, 
and let SPMD use a constant stepsize of 
\begin{align}\label{generic_muti_traj_stepsize}
\eta  = \tfrac{1}{2} \sqrt{\tfrac{\cD \mu}{k M^2} },
\end{align}
 where $M = \tfrac{1}{1-\gamma}$,
and  
$\cD > 0$ can be any constant satisfying 
\begin{align}\label{multi_traj_constant_D}
\textstyle
\cD \geq 2 \max_{s' \in \cS} \max_{\pi^* \in \overline{\Pi}^*} D^{\pi^*}_{\pi_0} (s') .
\end{align}
Let $\tau(\cD)$ be defined as in Condition \ref{divergence_condition_fixed_optimal}.
For any $\delta \in (0,1)$, let $Q^{\pi_t, \xi_t}$ be constructed by the TOMC operator defined in \eqref{eq_truncate_q},  with parameters $(m, n, \tau)$ specified as
 \begin{align}\label{eq_thrm_multi_exploration_m_n}
  \tau = \min \cbr{\tau(\cD), 1/\abs{\cA}}, 
 m = \Theta \rbr{\tfrac{\abs{\cA}^2}{(1-\gamma)^2 \varepsilon^2} \log (\tfrac{2 \abs{\cZ} k }{\delta}) },
n = \tilde{\Theta} \rbr{
\rbr{ \tfrac{1}{1-\gamma} 
+ \tfrac{\ceil{\log_\rho ( {\underline{\nu} \cdot {\tau}}/{(2 C)})}}{\underline{\nu}\cdot {\tau}} } \log (\tfrac{\abs{\cA} }{\varepsilon})
}, 
 \end{align}
 where $ \varepsilon = \tfrac{M}{2} \sqrt{\tfrac{\cD}{\mu k }}$.
Then with probability $1-\delta$, SPMD satisfies 
\begin{align*}
f(\hat{\pi}_k) - f(\pi^*)
\leq 
\tfrac{3 M }{1-\gamma} \sqrt{ \tfrac{\cD}{\mu k}},
\end{align*}
where $f(\hat{\pi}_k) = \min_{0 \leq t \leq k-1} f(\pi_t)$. 
In addition, to attain $f(\hat{\pi}_k) - f(\pi^*) \leq \epsilon$, the total number of samples required by SPMD is bounded by 
\begin{align*}
\tilde{\cO} \rbr{
\tfrac{\cD \abs{\cA}^2}{(1-\gamma)^6 \mu \epsilon^4}
\rbr{
\tfrac{1}{1-\gamma} +  \tfrac{\ceil{\log_\rho ( {\underline{\nu} \cdot {\tau}}/{(2 C)})}}{\underline{\nu}\cdot {\tau}}}
\log (\tfrac{18 \abs{\cZ} M^2 \cD}{\delta (1-\gamma)^2 \mu \epsilon^2})
\log (\tfrac{\abs{\cA}}{\epsilon})
}.
\end{align*}
\end{theorem}

\begin{proof}
Fixing a deterministic optimal policy $\pi^* \in \overline{\Pi}^*$,
we begin by noting that condition \eqref{induction_stepsize_condition} in Lemma \ref{lemma_induction_bias_bound} holds with the choice of $\eta$ and $\varepsilon$. 
Indeed, 
direct calculation~yields 
\begin{align*}
 \tfrac{\eta^2 M^2 t }{2 \mu}  \leq   \cD/8,
 ~
 \eta t \varepsilon \leq  \cD/4.
\end{align*}
Combining the above observation with the choice of $(m, n)$ specified in \eqref{choice_m_n_general},   one can apply Lemma \ref{lemma_induction_bias_bound} and obtain that, 
with probability at least $(1-\delta)^{k}$, 
\begin{align}
& \eta (1-\gamma) \rbr{f(\pi_t) - f(\pi^*)}  \nonumber \\
  \overset{(a)}{=} & \eta \EE_{s \sim \vartheta} \EE_{s' \sim d_{s}^{\pi^*}} \inner{Q^{\pi_t}(s', \cdot)}{\pi_t(\cdot|s') - \pi^*(\cdot|s')}  \nonumber  \\
\overset{(b)}{ \leq} &    \EE_{s \sim d_{\vartheta}^{\pi^*}}  D^{\pi^*}_{\pi_t} (s) -    \EE_{s \sim d_{\vartheta}^{\pi^*}}  D^{\pi^*}_{\pi_{t+1}} (s)
+ \tfrac{\eta^2 M^2}{2 \mu} 
+  \eta \EE_{s \sim d_{\vartheta}^{\pi^*}}   \inner{\delta_t(s, \cdot)}{\pi_t(\cdot |s) - \pi^*(\cdot|s)}  \label{multi_traj_accumulated_noise} \\
\overset{(c)}{\leq} &
  \EE_{s \sim d_{\vartheta}^{\pi^*}}  D^{\pi^*}_{\pi_t} (s) -    \EE_{s \sim d_{\vartheta}^{\pi^*}}  D^{\pi^*}_{\pi_{t+1}} (s)
+ \tfrac{\eta^2 M^2}{2 \mu} 
+  \eta \varepsilon , \nonumber
\end{align}
for all $t \leq k-1$,
where $(a)$ uses Lemma \ref{lemma:performance_diff}, 
$(b)$ follows from after taking the expectation w.r.t. $s \sim d_{\vartheta}^{\pi^*}$ in \eqref{ineq_nonsmooth_vi} of Lemma \ref{lemma:three_point},  
and $(c)$ uses Lemma \ref{lemma_induction_bias_bound}.
Using $(1-\delta)^k \geq 1-k \delta$, we make substitution $\delta \to \delta/k$, and conclude that the probability of attaining the prior relation is at least $1-\delta$, with the choice of $(m,n)$ specified in \eqref{eq_thrm_multi_exploration_m_n}.

Summing up the above relation from $t=0$ to $k-1$, 
and using $\cD \geq 2 \max_{s' \in \cS, \pi^* \in \overline{\Pi}^*} D^{\pi^*}_{\pi_0} (s') $,  
we obtain 
\begin{align*}
f(\hat{\pi}_k) - f(\pi^*)
\leq 
\tfrac{\cD }{2 \eta (1-\gamma) k}
+ \tfrac{\varepsilon}{1-\gamma} 
+ \tfrac{\eta M^2}{2 \mu (1-\gamma)},
\end{align*}
where $f(\hat{\pi}_k) = \min_{0 \leq t \leq k-1} f(\pi_t)$.
Substituting $\eta = \tfrac{1}{2} \sqrt{\tfrac{\cD \mu}{k M^2} }$ into the above relation yields 
\begin{align*}
f(\hat{\pi}_k) - f(\pi^*)
\leq 
\tfrac{2 M }{1-\gamma} \sqrt{ \tfrac{\cD}{\mu k}} + \tfrac{\varepsilon}{1-\gamma} 
\leq \tfrac{3 M }{1-\gamma} \sqrt{ \tfrac{\cD}{\mu k}} .
\end{align*}
To attain $f(\hat{\pi}_k) - f(\pi^*) \leq \epsilon$, it suffices to take $k = \frac{9 M^2 \cD}{\mu (1-\gamma)^2 \epsilon^2}$. The total number of samples is bounded by 
\begin{align*}
k \cdot m \cdot n
 & = \tilde{\Theta} \rbr{
  \frac{9 M^2 \cD}{\mu (1-\gamma)^2 \epsilon^2} \cdot
 \tfrac{\abs{\cA}^2}{(1-\gamma)^2 \varepsilon^2} \log (\tfrac{2 \abs{\cZ} k }{\delta}) \cdot
 \rbr{ \tfrac{1}{1-\gamma} 
+ \tfrac{\ceil{\log_\rho ( {\underline{\nu} \cdot {\tau}}/{(2 C)})}}{\underline{\nu}\cdot {\tau}} } \log (\tfrac{\abs{\cA} }{\varepsilon})
 } \\
 & = \tilde{\cO} \rbr{
 \tfrac{\cD \abs{\cA}^2}{(1-\gamma)^6 \mu \epsilon^4}
\rbr{
\tfrac{1}{1-\gamma} + \tfrac{\ceil{\log_\rho ( {\underline{\nu} \cdot {\tau}}/{(2 C)})}}{\underline{\nu}\cdot {\tau}}}
\log (\tfrac{18 \abs{\cZ} M^2 \cD}{\delta (1-\gamma)^2 \mu \epsilon^2})
\log (\tfrac{\abs{\cA}}{\epsilon})
 },
\end{align*}
where the second equality uses $M = \frac{1}{1-\gamma}$.
The proof is then completed. 
\end{proof}

In view of Theorem \ref{global_convergence_generic}, using multiple trajectories within TOMC incurs a total sample complexity of $\tilde{\cO}(1/\epsilon^4)$ for SPMD.
This already improves existing sample complexity bounds of stochastic PG methods that control the optimality gap in high probability without explicit exploration over the actions (e.g. \cite{hu2022actorcritic}).
In Section \ref{subsec_single_traj}, we will further introduce an improvement that obtains the optimal $\tilde{\cO}(1/\epsilon^2)$ sample complexity by using only a single trajectory in TOMC. 
Before that, let us consider a few applications of Theorem \ref{global_convergence_generic} to some concrete Bregman divergences applicable to policy optimization.



To specialize the global convergence of SPMD to any Bregman divergence of interest, 
it suffices to choose
the constant $\cD$ satisfying \eqref{multi_traj_constant_D}, verify Condition \ref{divergence_condition_fixed_optimal}, and consequently determine values of $(\mu, \tau(\cD))$ therein.
These parameters will then be used to determine $(m, n, \tau)$ of the TOMC operator specified in \eqref{eq_thrm_multi_exploration_m_n}, and the stepsize of SPMD specified in \eqref{generic_muti_traj_stepsize}. 
We first consider SPMD instantiated with the KL divergence.


\begin{proposition}[SPMD with KL Divergence]\label{prop_kl}
Suppose Assumption \ref{assump_unif_mixing} holds.
Fix total iterations $k > 0$, 
 and let SPMD
 adopt the KL divergence \eqref{eq_def_kl} with a constant stepsize $\eta  =  \sqrt{\tfrac{ \log \abs{\cA} }{2 k M^2} }$, where~$M = {1}/\rbr{1-\gamma}$.
For any $\delta \in (0,1)$, let $Q^{\pi_t, \xi_t}$ be constructed by the TOMC operator defined in \eqref{eq_truncate_q},  with parameters $(m, n, \tau)$ specified as
 \begin{align*}
\tau =  \abs{\cA}^{-2/(1-\gamma)},  ~
m = \Theta \rbr{\tfrac{\abs{\cA}^2}{(1-\gamma)^2 \varepsilon^2} \log (\tfrac{2 \abs{\cZ} k }{\delta}) },~
n = \tilde{\Theta} \rbr{
\rbr{
\tfrac{1}{1-\gamma} + \tfrac{ t_{\mathrm{mix}}}{\underline{\nu}} \abs{\cA}^{2/(1-\gamma)}} \log (\tfrac{\abs{\cA} }{\varepsilon})
}, 
 \end{align*}
 where $ \varepsilon =  M \sqrt{\tfrac{\log \abs{\cA} }{ 2  k }}$ 
 and $ t_{\mathrm{mix}} =  \ceil{\log_\rho(\underline{\nu}/(2C)) - 2 \log_\rho (\abs{\cA} ) / (1-\gamma) }$.
Then with probability $1-\delta$, we have 
\begin{align*}
f(\hat{\pi}_k) - f(\pi^*)
\leq 
\tfrac{3 M }{1-\gamma} \sqrt{ \tfrac{2 \log \abs{\cA}}{ k}},
\end{align*}
where $f(\hat{\pi}_k) = \min_{0 \leq t \leq k-1} f(\pi_t)$. 
In addition, to attain $f(\hat{\pi}_k) - f(\pi^*) \leq \epsilon$, the total number of samples required by SPMD is bounded by 
\begin{align*}
\tilde{\cO} \rbr{
\tfrac{ \abs{\cA}^2 \log \abs{\cA}}{(1-\gamma)^6  \epsilon^4}
\rbr{
\tfrac{1}{1-\gamma} + \tfrac{ t_{\mathrm{mix}}}{\underline{\nu}} \abs{\cA}^{2/(1-\gamma)}}
\log (\tfrac{36 \abs{\cZ}  \log \abs{\cA}}{\delta (1-\gamma)^4})
\log (\tfrac{\abs{\cA}}{\epsilon})
}.
\end{align*}
\end{proposition}

\begin{proof}
Given that $\pi_0$ is the uniform policy,  we have  $D^{\pi}_{\pi_0}(s) \leq \log \abs{\cA}$,
and hence one can choose 
$\cD = 2 \log \abs{\cA}$ so that \eqref{multi_traj_constant_D} is satisfied.
It remains to verify Condition \ref{divergence_condition_fixed_optimal} and determine $(\mu, \tau(\cD))$.

Part 2 of Condition \ref{divergence_condition_fixed_optimal} is trivial given \eqref{close_form_with_kl}.
We proceed to verify Parts 1 and 3 of Condition \ref{divergence_condition_fixed_optimal}.
From Pinsker's inequality,   $w(\cdot)$ is 1-strongly convex w.r.t $\norm{\cdot}_1$-norm, hence Part 1 of Condition \ref{divergence_condition_fixed_optimal} is satisfied with $\mu = 1$. 
On the other hand, for any $\pi^* \in \overline{\Pi}^*$,  $\pi \in \mathrm{ReInt}(\Pi)$, and  $s\in \cS$,   $(1-\gamma) D^{\pi^*}_{\pi} (s)  \leq   \cD$ is equivalent to 
\begin{align*}
D^{\pi^*}_{\pi}(s) = \tsum_{a \in \cA} \pi^*(a|s) \log (\tfrac{\pi^*(a|s)}{\pi(a|s)}) 
= - \log \pi(a^{\pi^*}_s |s) \leq \tfrac{2 \log \abs{\cA}}{1-\gamma},
\end{align*}
which implies $\pi(a^{\pi^*}_s|s) \geq \abs{\cA}^{-2/(1-\gamma)}$.
Hence one can take $\tau(\cD) = \abs{\cA}^{-2/(1-\gamma)}$ so that Part 3 of Condition \ref{divergence_condition_fixed_optimal} is satisfied.
Finally, plugging the choice of $(\mu, \cD, \tau(\cD))$ into Theorem \ref{global_convergence_generic} completes the proof.
\end{proof}

In view of Proposition \ref{prop_kl},  SPMD with the KL divergence yields an $\cO(\cH_{\cD} /\epsilon^4)$ sample complexity. 
On the other hand, it is also clear that $\cH_{\cD} = \Omega(\abs{\cA}^{-2/(1-\gamma)})$, which depends exponentially on size of the effective horizon $1/(1-\gamma)$, with the base being the size of the action space $\abs{\cA}$. 
We next introduce an alternative Bregman divergence that uses the negative Tsallis entropy as the distance-generating function, which substantially improves the pessimistic dependence on $\abs{\cA}$ and ${1}/(1-\gamma)$.

\begin{definition}[Tsallis Divergence]
Let the distance-generating function in \eqref{def:kl_bregman} be chosen as
\begin{align*}
w(\pi(\cdot|s)) = \tsum_{a \in \cA} - \pi(a|s)^p
\end{align*}
 for some $p \in (0,1)$.
That is, $w(\cdot)$ corresponds to the negative Tsalli entropy with an entropic-index $p$, up to a constant factor.
Then the induced Bregman divergence defined  in \eqref{def:kl_bregman} admits the following form
\begin{align}\label{eq_tsallis_divergence}
D^{\pi}_{\pi'}(s) =
\tsum_{a \in \cA} - (\pi(a|s))^p + (1-p) \tsum_{a \in \cA} (\pi'(a|s))^p
+ p \tsum_{a \in \cA} \pi(a|s) (\pi'(a|s))^{p-1}.
\end{align}
We refer to the divergence above as the Tsallis divergence with index $p \in (0,1)$.
\end{definition}

Incorporating Tsallis entropy into policy optimization has been discussed in \cite{zhan2021policy, lee2019tsallis}.
On the other~hand, the methods therein directly add the Tsallis entropy into the cost function, and focus on solving the regularized MDP. 
There seems to be no explicit discussion on incorporating the Tsallis divergence defined above into the policy update, for solving the original, unregularized MDP.

On the other hand, it should also be noted that the policy update step \eqref{spmd_update} of SPMD with Tsallis divergence does not have a closed-form solution. 
We next propose a simple bisection-based subroutine that can solve the policy update step to any accuracy in logarithmic number of steps.

\begin{proposition}[Efficient Policy Update with Tsallis Divergence]\label{prop_tsallis_update}
Consider SPMD update \eqref{spmd_update} with the Tsallis divergence with index $p \in (0,1)$. That is,
\begin{align}
\textstyle
\label{spmd_update_tsalli}
\pi_{k+1}(\cdot | s) = \argmin_{p(\cdot |s) \in \Delta_{\cA}} \eta_k \inner{Q^{\pi_k, \xi_k}(s, \cdot)}{p(\cdot|s)}   -
\tsum_{a \in \cA}  (\pi(a|s))^p
+ p \tsum_{a \in \cA} \pi(a|s) (\pi_k(a|s))^{p-1}.
\end{align}
 Define univariate functions
$\psi_a(\mu) \coloneqq (\tfrac{p}{q_a - \mu})^{1/(1-p)}$, and 
$\phi(\mu) \coloneqq \tsum_{a \in \cA}  \psi_a(\mu) - 1$, with $q_a \coloneqq \eta_k Q^{\pi_k, \xi_k}(s, a) + p \cdot (\pi_k(a|s))^{p-1}$ for every $a \in \cA$.
In addition, let $l \coloneqq \min_{a \in \cA} q_a - p \abs{\cA}^{1-p}$ and $h \coloneqq \min_{a \in \cA} q_a - p$.
Then
\begin{align}\label{property_root_function}
\phi ~\text{is strictly increasing on}~ [l, h], ~\text{and has a unique root}~ \mu^* \in [l, h].
\end{align}

Let $\hat{\mu}^*$ be the output from the standard root-finding bisection method 
\cite{ben2012optimization}
 applied to $\phi$ on $[l,h]$ for $B$ steps,
and consequently define $\overline{x} \in \RR^{\abs{\cA}}$ as
$
\overline{x}_a =  \psi_a(\hat{\mu}^*)  / \tsum_{a'}  \psi_{a'}(\hat{\mu}^*)
$
for every $a \in \cA$.
Then for any $\epsilon \in (0,1)$, 
\begin{align}\label{tsallis_subproblem_num_steps}
\overline{x} \in \mathrm{ReInt}(\Delta_{\cA}), ~ \text{and}~ \norm{\overline{x} - \pi_{k+1}(\cdot|s)}_\infty \leq \epsilon,
\end{align}
provided $B \geq \ceil{\log_2 (\tfrac{2 \abs{\cA}^2}{(1-p) \epsilon})}$.
\end{proposition}

The proof of Proposition \ref{prop_tsallis_update} is deferred to Appendix \ref{sec_supp}. 
We proceed to show that SPMD with the Tsallis divergence leads to a much improved dependence on the  size of the action space and the effective horizon. 
For simplicity of determining constants, we focus on the case where index  $p = 1/2$.

\begin{proposition}[SPMD with Tsallis Divergence]\label{prop_tsallis}
Suppose Assumption \ref{assump_unif_mixing} holds.
Fix total iterations $k > 0$, 
and  let SPMD
 adopt the Tsallis divergence with index $p = 1/2$ and  a constant stepsize $\eta  =  \sqrt{\tfrac{ \abs{\cA}^{1/2} - 1 }{8 \abs{\cA}  k M^2} }$, where $M = {1}/\rbr{1-\gamma}$.
For any $\delta \in (0,1)$, let $Q^{\pi_t, \xi_t}$ be constructed by the TOMC operator defined in \eqref{eq_truncate_q},  with parameters $(m, n, \tau)$ specified as
$\tau =  \rbr{{4}/\rbr{1-\gamma}}^{-2} \abs{\cA}^{-1}$, and 
\begin{align*}
m = \Theta \rbr{\tfrac{\abs{\cA}^2}{(1-\gamma)^2 \varepsilon^2} \log (\tfrac{2 \abs{\cZ} k }{\delta}) },
n = \tilde{\Theta} \rbr{
\rbr{
\tfrac{1}{1-\gamma} + \tfrac{ t_{\mathrm{mix}}}{\underline{\nu} (1-\gamma)^2} \abs{\cA}} \log (\tfrac{\abs{\cA} }{\varepsilon})
}, ~\text{with} ~ \varepsilon = M  \sqrt{\tfrac{ 2 \abs{\cA}(\abs{\cA}^{1/2} -1) }{ k }} ,
\end{align*}
  where $ t_{\mathrm{mix}} = \ceil{\log_\rho (\underline{\nu}  (1-\gamma)^2 / (32C \abs{\cA}))}$.
Then with probability $1-\delta$, we have 
\begin{align*}
f(\hat{\pi}_k) - f(\pi^*)
\leq 
\tfrac{12 M }{1-\gamma} \sqrt{ \tfrac{\abs{\cA}^{3/2}}{ k}},
\end{align*}
where $f(\hat{\pi}_k) = \min_{0 \leq t \leq k-1} f(\pi_t)$. 
In addition, to attain $f(\hat{\pi}_k) - f(\pi^*) \leq \epsilon$, the total number of samples required by SPMD is bounded by 
\begin{align*}
\tilde{\cO} \rbr{
\tfrac{ \abs{\cA}^{7/2}}{(1-\gamma)^8  \epsilon^4}
\rbr{
\tfrac{1}{1-\gamma} + \tfrac{  t_{\mathrm{mix}} \cdot \abs{\cA}  }{\underline{\nu} (1-\gamma)^2}}
\log (\tfrac{288 \abs{\cZ}  \abs{\cA}^{3/2}}{\delta (1-\gamma)^4  \epsilon^2})
\log (\tfrac{\abs{\cA}}{\epsilon})
}.
\end{align*}

\end{proposition}

\begin{proof}

Consider Tsallis divergence with index $p \in (0,1)$.
For any $\pi^* \in \overline{\Pi}^*$,  we obtain from \eqref{eq_tsallis_divergence} that
\begin{align*}
D^{\pi^*}_{\pi_0} (s) =
 -1 + (1-p) \tsum_{a \in \cA} \abs{\cA}^{-p} + p \abs{\cA}^{-(p-1)} 
= -1 + \abs{\cA}^{1-p}.
\end{align*}
One can then take, with an additional subscript $p$ indicating the dependence on the index $p$, that $\cD_p = 2 ( \abs{\cA}^{1-p} - 1)$, for which \eqref{multi_traj_constant_D} is satisfied. 
We next verify Condition \ref{divergence_condition_fixed_optimal} and determine $(\mu_p, \tau(\cD_p))$.

Consider the SPMD update \eqref{spmd_update}, which now admits the following form,
\begin{align*}
\textstyle
\pi_{k+1}(\cdot | s) = \argmin_{p(\cdot |s) \in \Delta_{\cA}} \eta_k \inner{Q^{\pi_k, \xi_k}(s, \cdot)}{p(\cdot|s)}   -
\tsum_{a \in \cA}  (\pi(a|s))^p
+ p \tsum_{a \in \cA} \pi(a|s) (\pi_k(a|s))^{p-1}.
\end{align*}
Since $p < 1$, the subdifferential of the above objective is empty at any point $x \in \mathrm{ReBd}(\Delta_{\cA})$.
Consequently, $\pi_{k+1}(\cdot|s) \in \mathrm{ReInt}(\Delta_{\cA} )$ and Part 2 of Condition \ref{divergence_condition_fixed_optimal} is satisfied. 
We proceed to verify Parts 1 and 3 of Condition \ref{divergence_condition_fixed_optimal}.
It is clear that for any $\pi^* \in \overline{\Pi}^*$,  $\pi \in \mathrm{ReInt}(\Pi)$, and  $s \in \cS$,   $(1-\gamma) D^{\pi^*}_{\pi} (s) \leq \cD_p$ implies 
\begin{align*}
(1-\gamma) ( p  \pi^{p-1}(a^{\pi^*}_s|s) -1) 
\leq 2 ( \abs{\cA}^{1-p} -1 ),
\end{align*}
which in turn yields $\pi(a^{\pi^*}_s |s) \geq (\tfrac{2}{(1-\gamma) p})^{1/(p-1)} \abs{\cA}^{-1}$, and one can take $\tau(\cD_p) = (\tfrac{2}{(1-\gamma) p})^{1/(p-1)} \abs{\cA}^{-1}$ so that Part 3 of Condition \ref{divergence_condition_fixed_optimal} is satisfied.
In addition, noting that $ - \tfrac{\partial^2}{\partial x^2} x^p  = p(1-p) x^{p-2} \geq p(1-p)$ for $x \in (0,1)$, we have
\begin{align*}
D^{\pi}_{\pi'}(s) \geq p(1-p) \norm{\pi(\cdot|s) - \pi'(\cdot|s)}_2^2 \geq \tfrac{p (1-p)}{\abs{\cA}} \norm{\pi(\cdot|s) - \pi'(\cdot|s)}_1^2,
\end{align*}
from which one can take $\mu_p = \tfrac{p (1-p)}{\abs{\cA}}$, and Part 1 of Condition \ref{divergence_condition_fixed_optimal} is satisfied.
Finally,  plugging the above choice of $(\mu_p, \cD_p, \tau(\cD_p))$ with $p = {1}/{2}$ into Theorem \ref{global_convergence_generic} completes the proof.
\end{proof}

Comparing Proposition \ref{prop_kl} and \ref{prop_tsallis}, it is clear that the choice of Bregman divergence can greatly impact the sample complexity of SPMD with the TOMC evaluation operator. 
In particular, using Tsallis divergence yields a significantly better sample complexity than that of KL divergence, in terms of its dependence on the size of action space and the effective horizon.
This appears to be a new quantitative observation on the role of Bregman divergences in affecting the efficiency of stochastic policy optimization~methods.

A few remarks are in order to conclude our discussions in this subsection.
First, SPMD with TOMC operator seems to enjoy wider applicability compared to the VBE operator as it is compatible to a more general class of Bregman divergence. 
Second, the obtained sample complexity of $\tilde{\cO}(\cH_\cD / \epsilon^4)$ exhibits a clear dependence on the Bregman divergence, and controls the optimality gap in high probability. 
On the other hand, using multiple trajectories in TOMC incurs a price of worse sample complexity compared to that of the VBE operator.
In the following subsection, we show that indeed using a single trajectory suffices to ensure the global convergence of SPMD, which substantially improves the sample complexity.

\subsection{SPMD with Single-trajectory TOMC}\label{subsec_single_traj}


Our discussion in Section \ref{subsec_exp_multiple_traj_analysis} uses multiple independent trajectories in the TOMC  operator \eqref{eq_truncate_q}.
The technical motivation mainly arises from the purpose of bounding the accumulated noise, namely, the summation of the last term in \eqref{multi_traj_accumulated_noise} across all iterations of SPMD. 
This was handled in Lemma \ref{lemma_induction_bias_bound},  by bounding the noise term at each iteration with high probability, consequently requiring the usage of multiple trajectories in the TOMC operator.

Apparently, a much more appealing option is to use a single trajectory for online policy evaluation. 
It turns out that using a single trajectory is indeed enough to control the accumulated noise.
Consequently, we are able to strengthen the sample complexity of SPMD with TOMC operator from $\tilde{\cO}(\cH_{\cD} /\epsilon^4)$ to $\tilde{\cO}(\cH_\cD /\epsilon^2)$.  
To this end, we first establish the following technical observation that may be of independent interest.

\begin{lemma}\label{lemma_bounding_noise_sudo_martingale}
Fix $k > 0$, a finite index set $\cI$,   $\delta \in (0,1)$,  and $b > 0$.
Suppose for some $Z, \varepsilon > 0$ with
\begin{align*}
Z \geq 2  \max \{\sqrt{k} \varepsilon, 2 b \sqrt{ \log (k \abs{\cI} /\delta)}\},
\end{align*} 
there is 
 a probability space  $\mathfrak{P} = (\Omega, \cF, \PP)$, 
  a total $I$ sequences  of random variables $\{ \cbr{X_{t,i}}_{t=0}^{k-1} \}_{i \in \cI}$ with $X_{t,i}: \Omega \to \RR$,  and filtration $\cbr{\cF_t}_{t=0}^{k} \subseteq \cF$,  such that   $-b \leq X_{t,i} \leq b$, $\cF_0 = \cbr{\emptyset, \Omega}$, $X_{t, i} \in \cF_{t+1}$, and 
\begin{align}\label{def_condition_small_expectation}
\EE \sbr{ X_{t,i} | \cF_t } \mathbbm{1}_{\cbr{\max_{i \in \cI} Y_{t,i} \leq Z \sqrt{t}}} \leq \varepsilon, ~ \forall i \in \cI, 0 \leq t \leq k-1, 
\end{align}
where $\cbr{Y_{t,i}}_{t=0}^k$ is defined as $Y_{0,i} \coloneqq 0$ and $Y_{t+1, i}  \coloneqq Y_{t, i}  + X_{t, i}$, for every $i \in \cI$.
Denote $[k] \coloneqq \cbr{0, \ldots, k}$, then the constructed sequences $\{ \cbr{Y_{t, i}}_{t=0}^k \}_{i \in \cI}$ satisfy 
\begin{align*}
\PP(Y_{t,i} \leq Z\sqrt{t}, \forall t \in [k], \forall i \in \cI) \geq 1-\delta.
\end{align*}
\end{lemma}

Lemma \ref{lemma_bounding_noise_sudo_martingale} can be viewed as a generalization of  the Azuma–Hoeffding inequality, but with the following essential difference. 
For each $i \in \cI$, the increments $\cbr{X_{t, i}}_{t}$ is not a (super)-martingale sequence, as its conditional expectation is close-to-zero only on certain desirable events of the history $\cF_t$.
The purpose of Lemma \ref{lemma_bounding_noise_sudo_martingale} is to establish that such desirable events indeed occur with a high probability.
As will be clear in the ensuing Lemma \ref{lemma_bounded_noise_single_traj}, by taking $\cI = \cS$ and  $X_{t,s} =  \EE_{s' \sim d_s^{\pi^*}} \inner{\delta_t(s', \cdot)}{\pi_t(\cdot |s') - \pi^*(\cdot|s')}$, Lemma \ref{lemma_bounding_noise_sudo_martingale} provides an effective tool to control the accumulated noise in SPMD.


\begin{proof}[Proof of Lemma \ref{lemma_bounding_noise_sudo_martingale}]
For every $i \in \cI$, 
define sequences $\cbr{X_{t,i}'}$ and  $\cbr{Y_{t,i}'}$ by $Y_{t, i}' = 0$ and  
\begin{align*}
X_{t,i}' \coloneqq X_{t,i} \mathbbm{1}_{\cbr{\max_{i \in \cI}Y_{t, i} \leq Z \sqrt{t}} }, ~ Y_{t+1, i}' \coloneqq Y_{t, i}' + X_{t, i}', ~ 0 \leq t \leq k-1.
\end{align*}
 By definition, we  have 
$Y_{t, i}, Y_{t,i}' \in \cF_t$, and
\begin{align}\
\EE \sbr{ Y_{t+1, i}' | \cF_t} 
= \EE \sbr{ Y_{t,i}' + X_{t, i}' | \cF_t} 
& = Y_{t, i}' + \EE \sbr{  X_{t, i} \mathbbm{1}_{\cbr{\max_{i \in \cI} Y_{t, i} \leq Z \sqrt{t}} } \big| \cF_t} \nonumber  \\
& \overset{(a)}{=} Y_{t, i}' + \EE \sbr{  X_{t, i}  | \cF_t} \mathbbm{1}_{\cbr{\max_{i \in \cI} Y_{t,i} \leq Z \sqrt{t}} } \nonumber \\
& \overset{(b)}{\leq} Y_{t, i}' + \varepsilon, \label{Y_prime_approx_martingale}
\end{align}
where $(a)$ follows from $\mathbbm{1}_{\cbr{\max_{i \in \cI} Y_{t, i} \leq Z \sqrt{t}} } \in \cF_t$, and $(b)$ follows from \eqref{def_condition_small_expectation}.
We next apply the standard arguments of Azuma's inequality, which gives, for any $i \in \cI$ and any $t > 0$,
\begin{align*}
\PP ( Y_{t, i}' \geq x) 
& \leq \min_{\lambda > 0} \exp(-\lambda x) \cdot  \EE \sbr{ \exp \rbr{ \tsum_{t' = 0}^{t-1} \lambda( Y_{t'+1, i} -  Y_{t', i} ) } } \\
& \leq \min_{\lambda > 0} \exp(-\lambda x) \cdot  \EE \sbr{ \exp  \rbr{   \tsum_{t' = 0}^{t-2} \lambda( Y_{t'+1, i} -  Y_{t', i} )}  \cdot  \EE \sbr{ \lambda( Y_{t, i} -  Y_{t-1, i} )   \big| \cF_{t-1}}} \\
& \overset{(c)}{\leq} \min_{\lambda > 0} \exp(-\lambda x) \cdot \EE \sbr{ \exp \rbr{  \tsum_{t' = 0}^{t-2} \lambda( Y_{t'+1, i} -  Y_{t', i} )  } }   \cdot \exp \rbr{\lambda \varepsilon + \tfrac{b^2 \lambda^2}{2}} \\
& \overset{(d)}{\leq} 
\min_{\lambda > 0} 
\exp(-\lambda x) 
\cdot 
\exp \rbr{
t \lambda \varepsilon + \tfrac{t b^2 \lambda^2}{2}
} \\
& = 
\exp \rbr{
\min_{\lambda > 0}
- (x - t \varepsilon) \lambda 
+ \tfrac{t b^2 \lambda^2}{2}
}
= \exp 
\rbr{
-\tfrac{(x - t \varepsilon)^2}{2 t b ^2}
},
\end{align*}
where $(c)$ follows directly from the Hoeffding's lemma combined with \eqref{Y_prime_approx_martingale}, and $(d)$ follows from a recursive application of inequality $(c)$.
Thus for any $\delta \in (0,1)$, by applying the union bound over $1 \leq t \leq k$ and $i \in \cI$ to the above relation,  we have 
\begin{align*}
Y_{t, i}' \leq t \varepsilon + 2 b \sqrt{\log(\tfrac{k \abs{\cI} }{\delta}) t}, ~ \forall t \in [k], ~ \forall i \in \cI , ~ \text{with probability at least}~ 1-\delta.
\end{align*}
Given the choice that $Z  \geq 2  \max \{\sqrt{k} \varepsilon, 2 b \sqrt{ \log ( \tfrac{k \abs{\cI} }{\delta})}\}$, we obtain 
\begin{align*}
Y_{t, i} ' \leq  Z \sqrt{t}, ~ \forall t \in [k],  ~ \forall i \in \cI, ~ \text{with probability at least}~ 1-\delta.
\end{align*}
Now define $\cG \coloneqq \{\omega: Y_{t, i}', \leq Z \sqrt{t}, ~\forall t \in [k], ~ \forall i \in \cI \}$, we proceed to show inductively that 
for any $\omega \in \cG$,
\begin{align}\label{inductive_claim}
Y_{t, i} (\omega) \leq Y_{t, i} '(\omega)  , ~ \forall i \in \cI,  ~ \forall t \in [k].
\end{align}
Note that $Y_{0, i} \equiv 0 \equiv Y_{0, i}'$, hence the claim \eqref{inductive_claim} holds at $t=0$.
Suppose \eqref{inductive_claim} holds at step $t$, then we have for every $\omega \in \cG$ and every $i \in \cI$, 
\begin{align*}
Y_{t+1, i}(\omega) = Y_{t, i}(\omega) + X_{t, i}(\omega) 
& \overset{(a')}{\leq} Y_{t, i}' (\omega) + X_{t, i}(\omega)   \\
& \overset{(b')}{\leq} Y_{t, i}' (\omega) +  X_{t, i}\mathbbm{1}_{\cbr{\max_{i \in \cI} Y_{t,i} \leq Z \sqrt{t}} } (\omega)
+ b \mathbbm{1}_{\cbr{\max_{i \in \cI} Y_{t, i} > Z \sqrt{t}} } (\omega) \\
& \overset{(c')}{=} Y_{t, i}' (\omega) +  X_{t, i}' \mathbbm{1}_{\cbr{\max_{i \in \cI} Y_{t,i} \leq Z \sqrt{t}} } (\omega)
+ b \mathbbm{1}_{\cbr{\max_{i \in \cI} Y_{t, i} > Z \sqrt{t}} } (\omega) \\
& \overset{(d')}{=} Y_{t, i}' (\omega)+  X_{t, i}'  (\omega)
= Y_{t+1, i}' (\omega),
\end{align*} 
where $(a')$ follows from the induction hypothesis that $Y_t(\omega) \leq Y_t'(\omega)$ for $\omega \in \cG$;
$(b')$ follows from $X_{t, i} (\omega) \leq b $ for any $\omega \in \Omega$; 
$(c')$ follows from $X_{t,i}'(\omega) = X_{t, i}(\omega)$ if $\max_{i \in \cI} Y_{t,i} (\omega) \leq  Z\sqrt{t}$, given the definition of $X_{t,i}'$.
Moreover, 
$(d')$ follows from that for $\omega \in \cG$, we have $\max_{i \in \cI} Y_{t,i} (\omega) \leq \max_{i \in \cI} Y_{t, i}' (\omega) \leq Z \sqrt{t}$, where the first inequality follows from the induction hypothesis, and the second inequality follows from the definition of $\cG$.
 Thus the induction is completed.

In summary, we have shown  $\PP(\cG ) \geq 1-\delta$. In addition,  for any $\omega \in \cG$,
$Y_{t, i} (\omega) \leq Y_{t, i}'(\omega) \leq Z \sqrt{t}$ holds for every $t  \in [k]$, $i \in \cI$. 
Hence $\PP(\cbr{\omega: ~Y_{t,i} \leq Z\sqrt{t}, \forall t \in [k], i \in \cI}) \geq 1-\delta$, and the proof is completed.
\end{proof}
%
%
%

With Lemma \ref{lemma_bounding_noise_sudo_martingale} in place, we proceed to establish the following bound on the accumulated noise in the SPMD method.

\begin{lemma}\label{lemma_bounded_noise_single_traj}
Suppose Assumption \ref{assump_unif_mixing} holds,
and the Bregman divergence \eqref{def:kl_bregman} satisfies Condition \ref{divergence_condition_fixed_optimal}.
Fix total iterations $k > 0$, $\delta \in (0,1)$. 
Set $b = \tfrac{1}{1-\gamma}$, 
and suppose $Z, \varepsilon > 0$ satisfy
$ Z \geq 2  \max \{\sqrt{k} \varepsilon, 2 b \sqrt{ \log (k \abs{\cS} /\delta)}\}$.
Set $(m, n, \tau)$ in the construction \eqref{eq_truncate_q} of TOMC estimator as 
\begin{align*}
\tau = \min \cbr{\tau(\cD_Z), 1/\abs{\cA}}, m = 1, ~ n = \tilde{\Theta} \rbr{
\rbr{ \tfrac{1}{1-\gamma} 
+ \tfrac{\ceil{\log_\rho ( {\underline{\nu} \cdot {\tau}}/{(2 C)})}}{\underline{\nu}\cdot {\tau}} } \log (\tfrac{\abs{\cA} }{\varepsilon})
},
\end{align*}
 where $\cD_Z > 0$ can be any constant satisfying 
 $\cD_Z \geq (2 + Z)    \max_{s \in \cS, \pi^* \in \overline{\Pi}^*} D^{\pi^*}_{\pi_0} (s)$.
In addition, set the stepsize  $ \eta_t = \eta > 0$ with
\begin{align}\label{single_traj_stepsize}
 \tfrac{\eta^2 M^2 k}{2 \mu} \leq  \tfrac{\cD_Z}{2+Z}, 
~
\eta \sqrt{k} \leq   \tfrac{\cD_Z}{2+Z} ,
\end{align}
where $M = \frac{1}{1-\gamma}$.
 Then for any deterministic optimal policy $\pi^* \in \overline{\Pi}^*$, 
\begin{align*}
\tsum_{i=0}^{t-1} \EE_{s' \sim d_s^{\pi^*}}  \inner{\delta_i(s', \cdot)}{\pi_i(\cdot |s') - \pi^*(\cdot|s')} \leq Z \sqrt{t}, ~ \forall 0 \leq t \leq k, ~ \forall s \in \cS, 
\end{align*}
holds with probability at least $1-\delta$.
\end{lemma}

\begin{proof}
 Let us fix a deterministic optimal policy $\pi^* \in \overline{\Pi}^*$ for the remainder of the proof,
 and define $a_s^{\pi^*}$ to be the unique action satisfying $\pi^*(a_s^{\pi^*} | s) =1$.
 By
summing up \eqref{divergence_to_optimal_bound} from $i=0$ to $t-1$, and 
taking $\eta_t = \eta$ therein, we obtain 
\begin{align}
(1-\gamma) D^{\pi^*}_{\pi_t} (s) & \leq
\EE_{s' \sim d_s^{\pi^*}} D^{\pi^*}_{\pi_t} (s') \nonumber  \\
& \leq \EE_{s' \sim d_s^{\pi^*}} D^{\pi^*}_{\pi_0} (s')
+ \tfrac{\eta^2 M^2 t }{2 \mu} 
+  \eta \tsum_{i=0}^{t-1} \EE_{s' \sim d_s^{\pi^*}}  \inner{\delta_i(s', \cdot)}{\pi_i(\cdot |s') - \pi^*(\cdot|s')}, \label{divergence_to_opt_single}
\end{align} 
where \eqref{divergence_to_opt_single} follows from $\norm{Q^{\pi_t, \xi_t}}_\infty \leq M = \frac{1}{1-\gamma}$.

Fix $k > 0$, $\delta \in (0,1)$. 
Set $b = \tfrac{1}{1-\gamma}$, $\cI = \cS$, 
and let $(Z, \varepsilon)$ satisfy
$ Z \geq 2  \max \{\sqrt{k} \varepsilon, 2 b \sqrt{ \log (k \abs{\cS} /\delta)}\}$.
We proceed to construct the required  $(\mathfrak{P}, \cbr{\cF_t}, \cbr{X_{t, s}}, \cbr{Y_{t, s}})$ satisfying the conditions in Lemma \ref{lemma_bounding_noise_sudo_martingale}.
Fix $n$, 
 let $\Omega $ denote the set of all possible outcomes of  trajectories $\cbr{\xi_t}_{t=0}^{k-1}$ when running the SPMD method for $k$ iterations, where each $\xi_t$ is a trajectory of length $n$. 
 Let $\cF$ be the power set (i.e., discrete $\sigma$-algebra) of $\Omega$. 
 Accordingly,
we define, for every $t \geq 0$, $\cF_{t+1}$, as the $\sigma$-algebra generated by $\cbr{\xi_{i}}_{i \leq t}$, and 
$\cF_0 = \cbr{\emptyset, \Omega}$.

Let us define, for every $s \in \cS$,
\begin{align*}
X_{t,s} \coloneqq  \EE_{s' \sim d_s^{\pi^*}} \inner{\delta_t(s', \cdot)}{\pi_t(\cdot |s') - \pi^*(\cdot|s')},  ~Y_{0, s} \coloneqq 0, 
~\text{and}~ 
Y_{t+1, s} \coloneqq Y_{t,s} + X_{t,s}.
\end{align*}
Clearly, we have $X_{t, s} \in \cF_{t+1}$ for all $t \geq 0$.
In addition,  $\abs{X_{t,s} } \leq b$.
It remains to properly choose $(n, \tau)$ so that \eqref{def_condition_small_expectation} holds.

 Suppose $\omega \in \Omega$ satisfies $\max_{s \in \cS} Y_{t,s}(\omega) \leq Z \sqrt{t}$,  then given the definition of $\cbr{ Y_{t,s}}$ and \eqref{divergence_to_opt_single}, we have
\begin{align*}
\textstyle
(1-\gamma) D^{\pi^*}_{\pi_t} (s) 
\leq \max_{s \in \cS}  D^{\pi^*}_{\pi_0} (s)
+ \tfrac{\eta^2 M^2 t }{2 \mu}
+ Z \eta \sqrt{t}, ~ \forall s \in \cS.
\end{align*}
Combining the above relation with the requirement of stepsize $\eta$ in \eqref{single_traj_stepsize} and $\cD_Z$, we further  obtain 
\begin{align*}
(1-\gamma) D^{\pi^*}_{\pi_t} (s) 
\leq  \cD_Z, ~ \forall s \in \cS.
\end{align*}
Thus given Condition \ref{divergence_condition_fixed_optimal},  there exists $\tau(\cD_Z) > 0$ such that 
\begin{align}\label{single_traj_prob_lb}
\pi_t(a^{\pi^*}_s |s) \geq \tau(\cD_Z), ~ \forall s \in \cS.
\end{align}
Now by taking $\tau = \min \cbr{\tau(\cD_Z), 1/\abs{\cA}}$ in the construction of $Q^{\pi_t, \xi_t}$ defined in \eqref{eq_truncate_q},  we obtain that, for every $s \in \cS$, 
\begin{align}
 &\EE \sbr{ \inner{\delta_t(s, \cdot)}{\pi_t(\cdot |s) - \pi^*(\cdot|s)}  |  \cF_t } \mathbbm{1}_{\cbr{\max_{s \in \cS} Y_{t,s} \leq Z \sqrt{t}}} \nonumber \\
\overset{(a)}{ \leq} &
2
\tsum_{a \in \cA, \pi_t(a|s) \geq \tau} \abs{ \EE \sbr{
 \delta_t(s,a)  | \cF_t
}}
+ \tsum_{\pi_t(a|s) < \tau}
\EE 
\sbr{
\delta_t(s,a)  \pi_t(a |s) 
  |  \cF_t
}  \mathbbm{1}_{\cbr{\max_{s \in \cS} Y_{t,s} \leq Z \sqrt{t}}} \nonumber\\
\overset{(b)}{\leq} & 2
\tsum_{a \in \cA, \pi_t(a|s) \geq \tau} \abs{ \EE \sbr{
 \delta_t(s,a)  | \cF_t
}} \nonumber  \\
\overset{(c)}{\leq} &
\tfrac{4\abs{\cA} (n+1)}{1-\gamma} \sbr{ \gamma^{n-1} + \rbr{1- \frac{\underline{\nu} \tau }{ \ceil{\log_\rho ( {\underline{\nu} \tau}/\rbr{2 C})} }}^{n-1}} .\label{ineq_from_y_to_each}
\end{align}
Here $(a)$ follows from \eqref{single_traj_prob_lb} and the definition of $\tau$, which states that any action $a$ with $\pi_t(a|s) < \tau$ must satisfy $a \neq a^{\pi^*}_s$,
and consequently $\pi^*(a|s) = 0$.
Inequality
$(b)$ follows from $\delta_t(s,a) \coloneqq Q^{\pi_t}(s,a) - Q^{\pi_t, \xi_t}(s,a) \leq 0$ for any action $a$ with $\pi_t(a|s) < \tau$, 
given the construction of $Q^{\pi_t, \xi_t}$ in \eqref{eq_truncate_q}, 
and hence the second term in $(a)$ can be dropped.
In addition, $(c)$  follows from applying 
Lemma \ref{lemma_bias_omc} to every action $a$ with $\pi_t(a|s) \geq \tau$.
Inequality \eqref{ineq_from_y_to_each} in turn implies that 
\begin{align*}
& \EE \sbr{ \EE_{s' \sim d_s^{\pi^*}}  \inner{\delta_t(s' , \cdot)}{\pi_t(\cdot |s') - \pi^*(\cdot|s' )}  |  \cF_t } \mathbbm{1}_{\cbr{\max_{s \in \cS} Y_{t,s} \leq Z \sqrt{t}}} \\
\leq &  \tfrac{4\abs{\cA} (n+1)}{1-\gamma} \sbr{ \gamma^{n-1} + \rbr{1- \frac{\underline{\nu} \tau }{ \ceil{\log_\rho ( {\underline{\nu} \tau}/\rbr{2 C})} }}^{n-1}}
, ~ \forall s \in \cS,
\end{align*}
which is equivalent to 
\begin{align*}
\EE \sbr{ X_{t,s} | \cF_t}   \mathbbm{1}_{\cbr{ \max_{s \in \cS} Y_{t,s} \leq Z \sqrt{t}}} 
\leq \tfrac{4\abs{\cA} (n+1)}{1-\gamma} \sbr{ \gamma^{n-1} + \rbr{1- \frac{\underline{\nu} \tau }{ \ceil{\log_\rho ( {\underline{\nu} \tau}/\rbr{2 C})} }}^{n-1}}, ~ \forall s \in \cS.
\end{align*}
By letting  
$
n = \tilde{\Theta} \rbr{
\rbr{ \tfrac{1}{1-\gamma} 
+ \tfrac{\ceil{\log_\rho ( {\underline{\nu} \tau}/{(2 C)})}}{\underline{\nu} \tau} } \log (\tfrac{\abs{\cA} }{\varepsilon})
},
$
it holds that $\EE \sbr{ X_{t, s} | \cF_t}   \mathbbm{1}_{\cbr{\max_{s \in \cS} Y_{t,s} \leq Z \sqrt{t}}}  \leq \varepsilon$ for any $s\in \cS$ and $\varepsilon > 0$, and \eqref{def_condition_small_expectation} is shown.

In conclusion, 
let the parameters $(m, n, \tau)$ defining the TOMC estimator $\cbr{Q^{\pi_t, \xi_t}}$ satisfy
\begin{align*}
m = 1, ~ n = \tilde{\Theta} \rbr{
\rbr{ \tfrac{1}{1-\gamma} 
+ \tfrac{\ceil{\log_\rho ( {\underline{\nu} \cdot {\tau}}/{(2 C)})}}{\underline{\nu}\cdot {\tau}} } \log (\tfrac{\abs{\cA} }{\varepsilon})
}, ~ \tau  = \min \cbr{\tau(\cD_Z), 1/\abs{\cA}},
\end{align*}
and the stepsize in SPMD satisfies \eqref{single_traj_stepsize},
then the constructed $(\mathfrak{P}, \cbr{X_{t,s}}, \cbr{Y_{t,s}}, \cbr{\cF_t})$ satisfy the conditions in Lemma \ref{lemma_bounding_noise_sudo_martingale}.
  Applying  Lemma \ref{lemma_bounding_noise_sudo_martingale} yields that 
\begin{align*}
\tsum_{i=0}^{t-1} \EE_{s' \sim d_s^{\pi^*}}  \inner{\delta_i(s', \cdot)}{\pi_i(\cdot |s') - \pi^*(\cdot|s')} \leq Z \sqrt{t}, ~ \forall 0 \leq t \leq k, ~ \forall s \in \cS, 
\end{align*}
holds probability at least $1-\delta$.
The proof is then completed.
\end{proof}

We are now ready to establish the generic convergence properties of SPMD, which uses the single-trajectory TOMC operator for policy evaluation.

\begin{theorem}\label{thrm_convergence_single_traj}
Suppose Assumption \ref{assump_unif_mixing} holds,
and the Bregman divergence \eqref{def:kl_bregman} satisfies Condition \ref{divergence_condition_fixed_optimal}.
Fix total iterations $k > 0$ a priori.
 For any $\delta \in (0,1)$, define 
\begin{align}\label{general_convergence_z_ve_choice}
Z=  \tfrac{4 \sqrt{ \log (\abs{\cS} k/\delta)} }{ 1-\gamma},
~ 
\varepsilon = 
\tfrac{2}{1-\gamma}  \sqrt{ \tfrac{ \log (\abs{\cS} k /\delta)}{ k}}.
\end{align}
Set the parameters $(m, n, \tau)$ constructing the TOMC estimator  \eqref{eq_truncate_q} as 
\begin{align}\label{sing_traj_m_n_tau_thrm}
m = 1, ~ n = \tilde{\Theta} \rbr{
\rbr{ \tfrac{1}{1-\gamma} 
+ \tfrac{\ceil{\log_\rho ( {\underline{\nu} \cdot \tau}/{(2 C)})}}{\underline{\nu}\cdot \tau} } \log (\tfrac{\abs{\cA} }{\varepsilon})
}, ~ \tau = \min \cbr{\tau(\cD_Z), 1/\abs{\cA}},
\end{align}
where 
 $\tau(\cD_Z)$ is defined as in Condition \ref{divergence_condition_fixed_optimal},
and $\cD_Z > 0$ can be any constant satisfying 
\begin{align}\label{ineq_DZ_single_traj}
\textstyle
\cD_Z \geq (2 + Z)    \max_{s \in \cS, \pi^* \in \overline{\Pi}^*} D^{\pi^*}_{\pi_0} (s).
\end{align}
In addition, set the stepsize  $\eta_t = \eta > 0$ with
\begin{align}\label{single_traj_stepsize_thrm}
\eta =  \min \cbr{
\sqrt{ \tfrac{2\cD_Z \mu}{(2+Z) M^2 }},
\tfrac{\cD_Z}{(2+Z) } 
} / \sqrt{k}.
\end{align}
Then  with probability at least $1- \delta$, we have
\begin{align}\label{ineq_single_traj_opt_gap_general}
f(\hat{\pi}_k) - f(\pi^*)
\leq 
\tfrac{\cD_Z }{(2+Z) \zeta^* (1-\gamma) \sqrt{k}}
+ \tfrac{\zeta^* M^2}{2 \mu (1-\gamma) \sqrt{k}}
+ \tfrac{Z}{(1-\gamma) \sqrt{k}}.
\end{align}
where $f(\hat{\pi}_k) = \min_{0 \leq t \leq k-1} f(\pi_t)$,  
$ 
\zeta^* = \min \cbr{
\sqrt{ \tfrac{2\cD_Z \mu}{(2+Z) M^2 }},
\tfrac{\cD_Z}{(2+Z) } 
}
$,
and $M = \frac{1}{1-\gamma}$.
To attain $f(\hat{\pi}_k) - f(\pi^*) \leq \epsilon$,
 the total number of iterations is bounded by 
\begin{align*}
k =
\Theta
\rbr{
\rbr{
\tfrac{Z^2}{(1-\gamma)^2}
+ 
\tfrac{M^4 (\zeta^*)^2 }{4 \mu^2 (1-\gamma)^2}
+ 
\tfrac{\cD_Z^2}{(2+Z)^2 (\zeta^*)^2 (1-\gamma)^2}
}\tfrac{1}{\epsilon^2}
}, 
\end{align*}
and the total number of samples is bounded by 
\begin{align*}
 \tilde{\Theta} \rbr{
\rbr{ \tfrac{1}{1-\gamma} 
+ \tfrac{\ceil{\log_\rho ( {\underline{\nu} \cdot \tau}/{(2 C)})}}{\underline{\nu}\cdot \tau} } 
\rbr{
\tfrac{Z^2}{(1-\gamma)^2}
+ 
\tfrac{M^4 (\zeta^*)^2 }{4 \mu^2 (1-\gamma)^2}
+ 
\tfrac{\cD_Z^2}{(2+Z)^2 (\zeta^*)^2 (1-\gamma)^2}
}\tfrac{1}{\epsilon^2}
}.
\end{align*}
\end{theorem}

\begin{proof}
Fix $k, \delta > 0$.
Note that $(Z, \varepsilon)$ specified in \eqref{general_convergence_z_ve_choice} satisfies 
$
Z \geq 2  \max \cbr{\sqrt{k} \varepsilon,  2b  \sqrt{ \log (\abs{\cS} k/\delta)} }$, 
where $b = \tfrac{1}{1-\gamma}$.
Suppose further that SPMD adopts a constant stepsize $\eta_t = \eta$ satisfying \eqref{single_traj_stepsize}.
Set the parameters $(n, m, \tau)$ of the TOMC operator as in \eqref{sing_traj_m_n_tau_thrm}, 
with $\cD_Z $ satisfying \eqref{ineq_DZ_single_traj}.
It is clear that all conditions in Lemma \ref{lemma_bounded_noise_single_traj} hold.
Fixing  $\pi^* \in \overline{\Pi}^*$, summing up \eqref{divergence_to_optimal_bound}  from $t=0$ to $k-1$ with $\eta_t = \eta$,  
and making use of $\norm{Q^{\pi_t, \xi_t}}_\infty \leq M = \tfrac{1}{1-\gamma}$,
we obtain 
\begin{align*}
& \eta ( 1-\gamma) \tsum_{t=0}^{k-1}  \sbr{V^{\pi_t}(s) - V^{\pi^*}(s)}   \\
 \leq &  \EE_{s' \sim d_s^{\pi^*}} D^{\pi^*}_{\pi_0} (s') 
+\tfrac{\eta^2 M^2 k}{2 \mu} 
+ \eta \tsum_{t = 0}^{k-1} \EE_{s' \sim d_s^{\pi^*}}  \inner{\delta_t(s', \cdot)}{\pi_t(\cdot |s') - \pi^*(\cdot|s')}  \\
\overset{(a)}{ \leq} &
\max_{s \in \cS} D^{\pi^*}_{\pi_0} (s) 
+ \tfrac{\eta^2 M^2 k}{2 \mu}
+ \eta Z \sqrt{k}, ~\forall s \in \cS, 
\end{align*}
with probability at least  $1-\delta$,
where $(a)$ applies Lemma \ref{lemma_bounded_noise_single_traj}.

Now further taking expectation of the above relation with respect to $s \sim \vartheta$, we obtain that,
 with probability at least $1-\delta$, 
\begin{align*}
& \eta ( 1-\gamma) \tsum_{t=0}^{k-1} \rbr{f(\pi_t) - f(\pi^*)}  \leq \tfrac{\cD_Z}{2+Z}
+ \tfrac{\eta^2 M^2 k}{2 \mu}
+ \eta Z \sqrt{k},
\end{align*}
or equivalently, 
\begin{align}
f(\hat{\pi}_k) - f(\pi^*)
\leq 
\tfrac{\cD_Z }{(2+Z) \eta (1-\gamma) k}
+ \tfrac{\eta M^2}{2 \mu (1-\gamma)}
+ \tfrac{Z}{(1-\gamma) \sqrt{k}} , \label{single_traj_opt_gap_before_opt_eta}
\end{align}
where $f(\hat{\pi}_k) = \min_{0 \leq t \leq k-1} f(\pi_t)$.
By minimizing the right hand side of \eqref{single_traj_opt_gap_before_opt_eta} under the constraint of $\eta > 0$ and \eqref{single_traj_stepsize}, we obtain the choice of $\eta$ in \eqref{single_traj_stepsize_thrm},  which consequently implies \eqref{ineq_single_traj_opt_gap_general}. 
Thus for $\hat{\pi}_k$ to be an $\epsilon$-optimal policy, it suffices to take 
\begin{align*}
k =
\Theta
\rbr{
\rbr{
\tfrac{Z^2}{(1-\gamma)^2}
+ 
\tfrac{M^4 (\zeta^*)^2 }{4 \mu^2 (1-\gamma)^2}
+ 
\tfrac{\cD_Z^2}{(2+Z)^2 (\zeta^*)^2 (1-\gamma)^2}
}\tfrac{1}{\epsilon^2}
}.
\end{align*}
By plugging the choice of $(Z, \varepsilon)$ specified in \eqref{general_convergence_z_ve_choice}  into \eqref{sing_traj_m_n_tau_thrm}, and combining the above bound of total iterations $k$, we obtain that
the total number of samples  is bounded by 
\begin{align*}
m \cdot n \cdot k
& =  
\tilde{\Theta} \rbr{
\rbr{ \tfrac{1}{1-\gamma} 
+ \tfrac{\ceil{\log_\rho ( {\underline{\nu} \cdot \tau}/{(2 C)})}}{\underline{\nu}\cdot \tau} } \log (\tfrac{\abs{\cA} (1-\gamma) \sqrt{k} }{ 2 \sqrt{ \log(\abs{\cS} k /\delta)}})
}
\cdot k
\\
 & =  
 \tilde{\Theta} \rbr{
\rbr{ \tfrac{1}{1-\gamma} 
+ \tfrac{\ceil{\log_\rho ( {\underline{\nu} \cdot \tau}/{(2 C)})}}{\underline{\nu}\cdot \tau} } 
\rbr{
\tfrac{Z^2}{(1-\gamma)^2}
+ 
\tfrac{M^4 (\zeta^*)^2 }{4 \mu^2 (1-\gamma)^2}
+ 
\tfrac{\cD_Z^2}{(2+Z)^2 (\zeta^*)^2 (1-\gamma)^2}
}\tfrac{1}{\epsilon^2}
}.
\end{align*}
The proof is then completed.
\end{proof}

Theorem \ref{thrm_convergence_single_traj} substantially reduces the number of trajectories used by the TOMC operator, from $\cO(1/\epsilon^2)$ stated in Theorem \ref{global_convergence_generic}, to exactly one. 
This reduction consequently improves the total sample complexity of SPMD from $\tilde{\cO}(1/\epsilon^4)$ to $\tilde{\cO}(1/\epsilon^2)$. 
Despite using a single trajectory, the obtained control of the optimality gap still holds in high probability. 
This appears to be the first $\tilde{\cO}(1/\epsilon^2)$ sample complexity among online PG methods without any explicit exploration, while attaining a high probability bound on the optimality gap.

Below, we specialize the generic convergence properties of SPMD to concrete Bregman divergences. 
Similar to our discussions in Section \ref{subsec_exp_multiple_traj_analysis}, 
it suffices to choose
a constant $\cD_{\cZ}$ satisfying \eqref{ineq_DZ_single_traj}, verify Condition \ref{divergence_condition_fixed_optimal}, and consequently determine the values of $(\mu, \tau(\cD_\cZ))$.
These in turn decide the concrete parameters $(m, n, \tau)$ of TOMC operator specified in \eqref{sing_traj_m_n_tau_thrm}, and the stepsize of SPMD specified in \eqref{single_traj_stepsize_thrm}. 
We first consider SPMD instantiated with the KL divergence.


%
%
%


%

\begin{proposition}[SPMD with KL Divergence]\label{prop_kl_single_traj}
Suppose Assumption \ref{assump_unif_mixing} holds.
Let SPMD adopt the KL divergence, and 
fix the total iterations $k > 0$ a priori.
 For any $\delta \in (0,1)$, define 
\begin{align*}
Z=  \tfrac{4 \sqrt{ \log (\abs{\cS} k/\delta)} }{ 1-\gamma},
~ 
\varepsilon = 
\tfrac{2}{1-\gamma}  \sqrt{ \tfrac{ \log (\abs{\cS} k /\delta)}{ k}}.
\end{align*}
Set the parameters $(m, n, \tau)$ constructing the TOMC estimator  \eqref{eq_truncate_q} as 
\begin{align*}
 \tau = \abs{\cA}^{-(2+Z)/(1-\gamma)}, ~ m = 1, ~ n = \tilde{\Theta} \rbr{
\rbr{ \tfrac{1}{1-\gamma} 
+ \tfrac{\ceil{\log_\rho ( {\underline{\nu} \cdot \tau}/{(2 C)})}}{\underline{\nu}\cdot \tau} } \log (\tfrac{\abs{\cA} }{\varepsilon})
}.
\end{align*}
In addition, set the stepsize in SPMD  as
\begin{align*}
\eta_t = \eta \coloneqq  \min \big\{
\sqrt{ \tfrac{2 \log \abs{\cA} }{M^2 }},
\log \abs{\cA} 
\big\} / \sqrt{k}, ~ M = \tfrac{1}{1-\gamma}, ~ \forall t = 0, \ldots k-1.
\end{align*}
Then  with probability at least $1- \delta$, we have
\begin{align*}
f(\hat{\pi}_k) - f(\pi^*)
\leq 
\tfrac{ \log \abs{\cA} }{\zeta^* (1-\gamma) \sqrt{k}}
+ \tfrac{\zeta^* M^2}{2  (1-\gamma) \sqrt{k}}
+ \tfrac{Z}{(1-\gamma) \sqrt{k}}, 
~
\zeta^* = \min \cbr{
\sqrt{ 2 \log \abs{\cA} (1-\gamma)^2},
\log \abs{\cA}
},
\end{align*}
where $f(\hat{\pi}_k) = \min_{0 \leq t \leq k-1} f(\pi_t)$.
To attain $f(\hat{\pi}_k) - f(\pi^*) \leq \epsilon$,
the total number of samples is bounded by 
\begin{align*}
\tilde{\cO} \rbr{
\rbr{
\tfrac{1}{1-\gamma} + \tfrac{ t_{\mathrm{mix}}}{\underline{\nu}} \abs{\cA}^{(2+Z)/(1-\gamma)}}
\rbr{
\tfrac{Z^2}{(1-\gamma)^2}
+ 
\tfrac{M^4 (\zeta^*)^2 }{4 (1-\gamma)^2}
+ 
\tfrac{\log^2 \abs{\cA}}{  (\zeta^*)^2 (1-\gamma)^2}
}\tfrac{1}{\epsilon^2}
}
,
\end{align*}
where
$t_{\mathrm{mix}}= \ceil{ \log_\rho(\underline{\nu}/(2C)) - (2+Z) \log_\rho (\abs{\cA} ) / (1-\gamma) }$.
\end{proposition}

\begin{proof}
Following the same lines as in the proof of Proposition \ref{prop_kl}, one can check that the choice of $\cD_Z = (2+ Z) \log \abs{\cA}$ satisifes  \eqref{ineq_DZ_single_traj}.
It remains to verify Condition \ref{divergence_condition_fixed_optimal}, and determine $(\mu, \tau(\cD_Z))$.

Parts 1 and 2 of Condition \ref{divergence_condition_fixed_optimal} are 
already shown in the proof of Proposition \ref{prop_kl}, with $\mu = 1$.
On the other hand, for any $\pi^* \in \overline{\Pi}^*$,  $\pi \in \mathrm{ReInt}(\Pi)$, and  $s \in \cS$,     $(1-\gamma) D^{\pi^*}_{\pi} (s)  \leq   \cD_Z$ is equivalent to 
\begin{align*}
D^{\pi^*}_{\pi}(s) = \tsum_{a \in \cA} \pi^*(a|s) \log (\tfrac{\pi^*(a|s)}{\pi(a|s)}) 
= - \log \pi(a^{\pi^*}_s|s) \leq \tfrac{(2+Z) \log \abs{\cA}}{1-\gamma},
\end{align*}
which implies $\pi(a^{\pi^*}_s|s) \geq \abs{\cA}^{-(2+Z)/(1-\gamma)}$.
Hence one can take $\tau(\cD_Z) = \abs{\cA}^{-(2+Z)/(1-\gamma)}$ so that Part 3 of Condition \ref{divergence_condition_fixed_optimal} is satisfied.
Plugging the choice of $(\mu, \cD_Z, \tau(\cD_Z))$ into Theorem \ref{thrm_convergence_single_traj} completes the proof.
\end{proof}


In view of Proposition \ref{prop_kl_single_traj}, the KL divergence-based SPMD combined with the single-trajectory TOMC operator attains an $\tilde{\cO}(\cH_{\cD} / \epsilon^2)$ 
sample complexity.
On the flip side, the term $\cH_{\cD}$ for KL divergence  scales exponentially with respect to the effective horizon, with a base being the size of the action space.
Similar to our discussions in Section \ref{subsec_exp_multiple_traj_analysis}, we next show that Tsallis divergence leads to a much improved dependence.

\begin{proposition}[SPMD with Tsallis Divergence]\label{prop_tsallis_single_traj}
Suppose Assumption \ref{assump_unif_mixing} holds.
Let SPMD adopt the Tsallis divergence with index $p =1/2$, and 
fix the total iterations $k > 0$ a priori.
 For any $\delta \in (0,1)$, define 
\begin{align*}
Z=  \tfrac{4 \sqrt{ \log (\abs{\cS} k/\delta)} }{ 1-\gamma},
~ 
\varepsilon = 
\tfrac{2}{1-\gamma}  \sqrt{ \tfrac{ \log (\abs{\cS} k /\delta)}{ k}}.
\end{align*}
Set the parameters $(m, n, \tau)$ constructing the TOMC estimator  \eqref{eq_truncate_q} as 
\begin{align*}
 \tau  = \tfrac{(1-\gamma)^2}{(4 + 2Z)^2} \abs{\cA}^{-1}, ~ m = 1, ~ n = \tilde{\Theta} \rbr{
\rbr{ \tfrac{1}{1-\gamma} 
+ \tfrac{\ceil{\log_\rho ( {\underline{\nu} \cdot \tau}/{(2 C)})}}{\underline{\nu}\cdot \tau} } \log (\tfrac{\abs{\cA} }{\varepsilon})
}.
\end{align*}
In addition, set the stepsize in SPMD as
\begin{align*}
\eta_t = \eta \coloneqq   \sqrt{\tfrac{A^{1/2} - 1}{ 2 \abs{\cA} M^2 k}}, ~ M = \tfrac{1}{1-\gamma}, ~ \forall t = 0, \ldots k-1.
\end{align*}
Then  with probability at least $1- \delta$, we have
\begin{align*}
f(\hat{\pi}_k) - f(\pi^*) 
 \leq \tfrac{4M}{1-\gamma} \sqrt{ \tfrac{\abs{\cA}^{3/2}}{k} }
+ \tfrac{4 \sqrt{ \log (\abs{\cS} k/\delta)}}{(1-\gamma)^2 \sqrt{k}},
\end{align*}
where $f(\hat{\pi}_k) = \min_{0 \leq t \leq k-1} f(\pi_t)$.
To attain $f(\hat{\pi}_k) - f(\pi^*) \leq \epsilon$,
the total number of samples is bounded by 
\begin{align*}
& \tilde{\cO} \rbr{
\rbr{
\tfrac{1}{1-\gamma} + \tfrac{  t_{\mathrm{mix}} (2 + Z)^2  \abs{\cA}  }{\underline{\nu} (1-\gamma)^2}}
\rbr{
\tfrac{Z^2}{(1-\gamma)^2}
+ 
\tfrac{ \abs{\cA}^{3/2} }{   (1-\gamma)^4}
}\tfrac{1}{\epsilon^2} },
\end{align*}
where $ t_{\mathrm{mix}} = \ceil{\log_\rho ( 2 \underline{\nu}  (1-\gamma)^2 / ((2+Z)^2C \abs{\cA}))}$.
\end{proposition}

\begin{proof}
Consider Tsallis divergence with index $p \in (0,1)$.
Following the same lines as in the proof of Proposition \ref{prop_tsallis}, one can take,  with an additional subscript indicating the dependence on $p$, that $\cD_{Z, p} = (2+Z) ( \abs{\cA}^{1-p} - 1)$, for which \eqref{ineq_DZ_single_traj} is satisfied. 
We next verify Condition \ref{divergence_condition_fixed_optimal}, and determine $(\mu_p, \tau(\cD_{Z, p}))$.

Parts 1 and 2 of Condition \ref{divergence_condition_fixed_optimal} are already shown in the proof of Proposition \ref{prop_tsallis}, with $\mu_p = \frac{p(1-p)}{\abs{\cA}}$. 
Meanwhile,  
for any $\pi^* \in \overline{\Pi}^*$,  $\pi \in \mathrm{ReInt}(\Pi)$, and $s\in \cS$,
$(1-\gamma) D^{\pi^*}_{\pi} (s) \leq \cD_{Z,p}$ implies 
\begin{align*}
(1-\gamma) ( p  \pi^{p-1}(a^{\pi^*}_s|s) -1) 
\leq (2+Z) ( \abs{\cA}^{1-p} -1 ),
\end{align*}
which gives $\pi(a^{\pi^*}_s|s) \geq (\tfrac{2+Z}{(1-\gamma) p})^{1/(p-1)} \abs{\cA}^{-1}$.
Hence one can take $\tau(\cD_{Z,p}) = (\tfrac{2+Z}{(1-\gamma) p})^{1/(p-1)} \abs{\cA}^{-1}$ so that Part 3 of Condition \ref{divergence_condition_fixed_optimal} is satisfied.
Finally, plugging the choice of $(\mu_p, \cD_{Z,p}, \tau(\cD_{Z, p}))$ with $p = {1}/{2}$ into Theorem \ref{thrm_convergence_single_traj} completes the proof.
%
%
%
\end{proof}

%
%


In view of Proposition \ref{prop_tsallis_single_traj}, it is clear that for SPMD instantiated with the Tsallis divergence of index $1/2$, the obtained $\tilde{\cO}(\cH_\cD/ \epsilon^2)$ sample complexity has a polynomial dependence on both the size of the action space and the effective horizon, an exponential improvement compared to that of using the KL divergence.

To conclude our discussion in this section, we briefly compare and highlight the differences between the development in Section  \ref{sec_spmd_vbe} and \ref{sec_spmd_omc}.
Namely, the difference between SPMD with   the VBE operator (Theorem \ref{thrm_spmd_vbe}), and SPMD with  the TOMC operator (Theorem \ref{thrm_convergence_single_traj}).

It is clear that SPMD with the VBE operator circumvents the technicality for bounding the accumulated noise through a probabilistic argument (Lemmas \ref{lemma_bounding_noise_sudo_martingale} and \ref{lemma_bounded_noise_single_traj}), 
and thus admits a simpler analysis compared to that of the TOMC operator. 
Nevertheless, this relative simplicity comes with a price in its practicality. 
First, 
 the optimality gap for the TOMC operator holds in high probability,
while it holds only in expectation for the VBE operator.
Second, SPMD can be combined with the TOMC operator for a more general class of Bregman divergences,
 while VBE operator seems to be limited to the KL divergence.
 Third, with the TOMC operator one can also relax Assumption \ref{assump_unif_mixing}, while it is not clear whether similar relaxation holds for the VBE operator (cf. Remark \ref{remark_relax_assump}).
 Finally, we believe the analysis of TOMC operator can be extended to solving regularized MDPs, and obtain a high probability bound on the optimality gap of the last-iterate~policy.


\section{Concluding Remarks}\label{sec_disccusion}

%
%
%

This manuscript establishes the sample complexity of a first-order stochastic policy optimization method, named stochastic policy mirror descent (SPMD), 
with two online policy evaluation operators that do not require explicit exploration over actions.
SPMD with the first evaluation operator, named value-based estimation (VBE), tailors to the KL divergence, and  attains an $\tilde{\cO}(1/\epsilon^2)$ sample complexity with linear dependence on the size of the action space. 
SPMD with the second evaluation operator, truncated online Monte-Carlo (TOMC) estimation, exhibits inherent exploration, in the sense that every optimal action is chosen with a non-diminishing probability throughout the optimization process. 
As a consequence, 
using a single trajectory
suffices to attain an $\tilde{\cO}(\cH_{\cD}/\epsilon^2)$ sample complexity with high probability, where $\cH_{\cD}$ depends on the Bregman divergence, size of the action space, and the effective horizon.

We now discuss a few directions worthy of future investigation.
First, the developed result holds in the best-iterate sense, which coincides with the notion of regret \cite{azar2017minimax}.
An apparently more appealing alternative is to establish similar sample complexity when the method outputs the last-iterate policy.
A potential approach would be adding a proper strongly-convex regularization into the cost function, 
and adapting the analysis of this manuscript (in particular, Section \ref{sec_spmd_omc}) to solve the regularized MDP.
Second, it seems interesting to explicitly characterize the difference between myopic exploration and inherent exploration in their sample complexities, 
 as SPMD with TOMC operator does not require learning the actions once identified as non-optimal ones. 
 Third, we believe analyses in this manuscript can be potentially extended to policy optimization with linear function approximation. 
Finally, it is also rewarding to design simple, model-free methods that maintain the inherent exploration of SPMD over actions and perform efficient exploration over the state space, while attaining the optimal $\cO(1/\epsilon^2)$ sample complexity.


%
%
%

\bibliographystyle{plain}
{\small
\bibliography{references}
}

\appendix 

\section{Supplementary Proof}\label{sec_supp}

\begin{proof}[Proof of Proposition \ref{prop_tsallis_update}]
It is clear that  the policy update step \eqref{spmd_update_tsalli}  is equivalent to 
\begin{align}\label{tsallis_update_problem}
\textstyle
 \min_{x \in \RR^{\abs{\cA}}, x \geq \mathbf{0}}
\inner{q}{x} - \tsum_{a \in \cA} x_a^p, ~ \text{s.t.} ~ \tsum_{a \in \cA} x_a = 1,
\end{align}
where $q_a = \eta_k Q^{\pi_k, \xi_k}(s, a) + p \cdot (\pi_k(a|s))^{p-1}$ for every $a \in \cA$.
Noting that \eqref{tsallis_update_problem} satisfies the relaxed Slater condition, we know that
for an optimal solution $x^*$ to \eqref{tsallis_update_problem}, 
 there exists a  Lagrange multiplier $\mu^* \in \RR$, 
\begin{align*}
\textstyle
x^* \in \Argmin_{x \in \RR^{\abs{\cA}}, x\geq \mathbf{0}}
\inner{q}{x} - \tsum_{a \in \cA} x_a^p - \mu^*(\tsum_{a \in \cA} x_a - 1) .
\end{align*} 
The above problem is clearly separable, which implies 
\begin{align*}
\textstyle
x^*_a \in \Argmin_{x_a \in \RR, x_a \geq {0}}
q_a x_a -  x_a^p - \mu^* x_a  , ~ \forall a \in \cA.
\end{align*}
Since  $x_a^* \leq 1$, then one must have $\mu^* < \min_{a\in \cA} q_a$,  otherwise the above program as unbounded infimum with $x_a \to \infty$.
In addition, since $p < 1$, we clearly have $x_a^* > 0$, and thus the first-order necessary optimality condition of the above program gives 
\begin{align}\label{solution_primal_based_on_dual}
q_a - p (x_a^*)^{p-1} - \mu^* = 0 ~\Rightarrow ~
x_a^* = (\tfrac{q_a - \mu^*}{p})^{1/(p-1)}.
\end{align}
On the other hand, 
from the Karush-Kuhn-Tucker condition, 
we also know that any $x^*$ given by \eqref{solution_primal_based_on_dual} (with $\mu^* < \min_{a \in \cA} q_a$), while satisfying $\tsum_{a \in \cA} x_a^* = 1$, is an optimal solution to \eqref{tsallis_update_problem}.

Now  plugging \eqref{solution_primal_based_on_dual} into constraint of \eqref{tsallis_update_problem} gives 
$
\tsum_{a \in \cA} (\tfrac{p}{q_a - \mu^*})^{1/(1-p)} = 1.
$
This implies that there must exists $a \in \cA$, such that $ (\tfrac{p}{q_a - \mu^*})^{1/(1-p)} \geq \tfrac{1}{\abs{\cA}}$, 
and hence $\mu^* \geq q_a - p \abs{\cA}^{1-p} \geq \min_{a \in \cA} q_a - p \abs{\cA}^{1-p} $.
In addition, for every $a \in \cA$, we have 
$(\tfrac{p}{q_a - \mu^*})^{1/(1-p)} < 1$, and hence $\mu^* \leq \min_{a \in \cA} q_a - p$.
Given our prior discussions, we see that 
$\mu^* \in [l, h]$ with $l = \min_{a \in \cA} q_a - p \abs{\cA}^{1-p}$ and $h = \min_{a \in \cA} q_a - p$.
In summary, 
\begin{align*}
\phi(\mu) \equiv \tsum_{a \in \cA}  (\tfrac{p}{q_a - \mu})^{1/(1-p)} - 1 ~ \text{has a solution $\phi(\mu^*) = 0$ with} ~ \mu^* \in [l,  h].
\end{align*}

In addition, as $\phi'(\mu) = \sum_{a \in \cA} (q-1)  \tfrac{p^q}{(q_a - \mu)^{q+1}}  > 0$ for $\mu \in [l, h]$, where $q \coloneqq \tfrac{1}{1-p} > 1$,  $\phi$ is strictly increasing on $[l, h]$,  with a unique root $\mu^* \in [l, h]$.
Hence \eqref{property_root_function} is proved.
Applying the standard bisection method to $\phi$ on interval $[l, h]$ for $B$ iterations, we obtain $\hat{\mu}^* \in [l, h]$ with $\abs{\hat{\mu}^* - \mu^*} 
\leq (h - l) 2^{-B} 
\leq \abs{\cA} 2^{-B}$.

Now define $\psi_a(\mu) = (\tfrac{q_a - \mu}{p})^{1/(p-1)}$, and let 
$
\hat{x}_a = \psi_a(\hat{\mu}^*)  , ~ x_a^* = \psi_a(\mu^*).
$
Note that both $\hat{x_a}, x_a^* > 0$, as we have $\hat{\mu}^*, \mu^* \leq h \equiv \min_{a \in \cA} q_a - p$. 
Define $\delta_a = \hat{x}_a - x_a^*$,  then from the mean value theorem 
\begin{align*}
\textstyle
\abs{\delta_a} \leq \sup_{\mu \in [l, h]} \abs{\psi_a'(\mu)} \abs{\hat{\mu}^* - \mu^*}
\leq \sup_{\mu \in [l, h]}  (q-1) \tfrac{p^q}{(q_a - \mu)^{q+1}}  \abs{\hat{\mu}^* - \mu^*}
\leq \tfrac{1}{1-p}  \abs{\hat{\mu}^* - \mu^*},
\end{align*}
where in the last inequality we use again $ h \leq \min_{a \in \cA} q_a - p$.
Given the above observation, let $\overline{x}_a = \hat{x}_a / \sum_{a' \in \cA} \hat{x}_a > 0$, 
it is clear that $\overline{x} \in \mathrm{ReInt}(\Delta_{\cA})$.
We then have 
\begin{align*}
\abs{ \overline{x}_a - x_a^*}  
= \abs{
\frac{x_a^* + \delta_a}{\tsum_{a'} (x_{a'}^* + \delta_a)} 
-
\frac{x_a^* }{\tsum_{a'} x_{a'}^* }  
}
=\abs{  \tfrac{\delta_a - x_a^* \sum_{a' \in \cA} \delta_{a'}}{1 + \sum_{a'} \delta_{a'}} }
\leq \tfrac{\abs{\cA} \abs{\hat{\mu}^* - \mu^*} / (1-p)}{1 - \abs{\cA} \abs{\hat{\mu}^* - \mu^*} / (1-p)}
\leq \tfrac{2 \abs{\cA}}{1-p} \abs{\hat{\mu}^* - \mu^*},
\end{align*}
provided $ \abs{\hat{\mu}^* - \mu^*} \leq (1-p) / (2 \abs{\cA})$.
Now since $ \abs{\hat{\mu}^* - \mu^*} \leq \abs{\cA} 2^{-B}$, 
we conclude that for any $\epsilon \in (0,1)$, to find $\abs{\overline{x}_a - x_a} \leq \epsilon$, it suffices let the number of iterations in the bisection procedure $B$ to satisfy
$
\tfrac{2 \abs{\cA}}{1-p} \cdot \abs{\cA} 2 ^{-B} \leq \epsilon.
$
Taking $B = \ceil{\log_2 (\tfrac{2 \abs{\cA}^2}{(1-p) \epsilon})}$  completes the proof for \eqref{tsallis_subproblem_num_steps}.
\end{proof}

\end{document}